\documentclass[10pt,twocolumn,letterpaper]{article}

\usepackage[pagenumbers]{cvpr} 

\usepackage{multirow}
\usepackage{longtable}
\usepackage{etoolbox}
\usepackage{makecell}
\usepackage{amsmath,amssymb}
\usepackage{dsfont}

\definecolor{cvprblue}{rgb}{0.21,0.49,0.74}
\usepackage[pagebackref,breaklinks,colorlinks,allcolors=cvprblue]{hyperref}

\def\confName{CVPR}
\def\confYear{2027}

\title{RSLiveBench-UAV: A Continually Evolving Vision-Language Benchmark for Aerial Spatial Intelligence}

\author{
Yue Zhou$^{1}$, Jiawei Liu$^{1}$, Jue Chen$^{1}$,
Ziyang Gong$^{2}$, Haojia Wei$^{4}$, Jingxiang Li$^{5}$\\
Xue Jiang$^{2}$, Zhihang Zhong$^{2}$, Pai Peng$^{3}$,
Xue Yang*$^{2,3}$, Yansheng Li$^{6}$, Jonathan Li*$^{5}$\\[0.4em]
$^{1}$East China Normal University \quad
$^{2}$Shanghai Jiao Tong University\\
$^{3}$Cowarobot \quad
$^{4}$University of California, Los Angeles\\
$^{5}$University of Waterloo \quad
$^{6}$Wuhan University
}

\begin{document}
\maketitle
\begin{abstract}
Aerial spatial intelligence (ASI) requires models to understand objects and spatial relationships across changing viewpoints and scales. Evaluating these abilities as models evolve calls for fresh physical observations, yet producing new aerial spatial questions requires repeatedly establishing entity identities, geometric references, and cross-view correspondences. We introduce \textbf{RSLiveBench-UAV}, a continually evolving vision-language benchmark driven by recurring acquisition of real UAV imagery. Its update protocol admits images captured after the preceding release and delays answer disclosure. An entity-centered production workflow uses geometry-assisted association to link repeated observations of physical vehicles, reuse human-verified attributes, and combine them with view-specific annotations and spatial references for traceable question generation. All tasks use deterministic scoring without model judges. We evaluate 30 general-purpose, remote-sensing, and embodied models on 35,491 question records from ten scenes, covering 17 tasks across 2D perception, 3D perception, cross-view perception, and reasoning. All 18 tested embodied models outperform the three selected remote-sensing comparators in Overall, although the strongest embodied model scores 43.29 versus 59.88 for GPT-6-Astra. Fine-grained vehicle recognition remains challenging, and performance varies across heights and camera pitches. Retrospective scene-group comparisons and rolling-window replay characterize score-pattern consistency under scene renewal. RSLiveBench-UAV connects fresh physical observations, entity-centered answer production, and systematic model evaluation to support continual assessment of ASI. Benchmark will release at \href{https://rslivebench.space}{https://rslivebench.space}
\end{abstract}

\section{Introduction}
\label{sec:introduction}


Spatial intelligence is the ability to perceive, represent, and reason about objects, their physical properties, and their relationships in space. For embodied models, these capabilities support the spatial understanding needed for navigation and interaction. Recent benchmarks have advanced their evaluation through visual perception, spatial reasoning, and multi-image understanding~\cite{fu2024blink,yang2025thinking,yang2026mmsibench}. A central question is how well this spatial understanding extends to observation conditions that differ substantially in viewpoint and scale.

Aerial spatial intelligence (ASI) provides an important setting for examining this question. Unmanned aerial vehicles (UAVs) observe the world across varying altitudes and camera orientations, which change object scale, visible surfaces, occlusion, and projected spatial relationships. Models must distinguish objects from overhead views, estimate physical quantities from perspective-dependent evidence, associate instances across images, and reason between camera-centered and object-centered reference frames.

\begin{figure*}[!t]
\centering
\includegraphics[width=\textwidth]{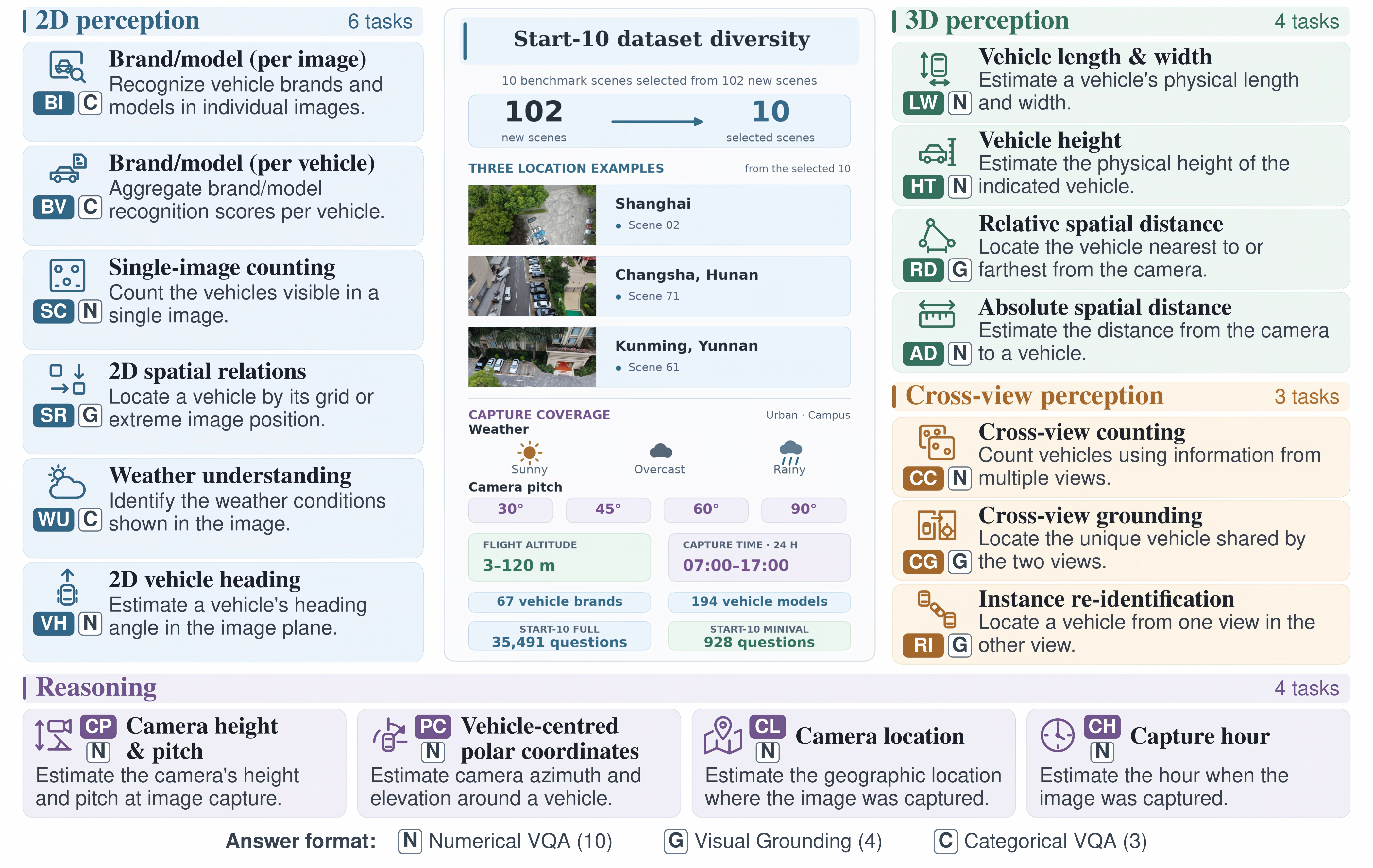}
\caption{Overview of the 17 benchmark tasks, grouped into 2D perception, 3D perception, cross-view perception, and reasoning.}
\label{fig:tasks}
\end{figure*}

Keeping this evaluation current as models evolve requires a continuing supply of fresh physical observations and the spatial references needed to assess them. This introduces two challenges. First, producing new spatial questions requires recurring field acquisition and renewed verification of object identities, physical attributes, camera geometry, and cross-view correspondences.  Second, replacing scenes changes the composition and difficulty of the evaluation set. Sustainable ASI evaluation therefore requires both a reusable answer-production workflow and an explicit renewal protocol supported by consistency analysis.

We introduce \textbf{RSLiveBench-UAV}, a continually evolving vision-language benchmark driven by recurring field acquisition of real UAV imagery, with an inventory of 102 collected scenes. Focusing initially on vehicles, our \textbf{entity-centered answer-production workflow} uses geometry-assisted association to link observations of the same vehicle within each capture batch and reuses verified make/model attributes to provide specification-based references across views. Templates combine entity records, view-specific annotations, and spatial references to generate traceable answers under human review. The live protocol replaces one scene monthly within a four-scene window, admitting only images captured after the preceding leaderboard release. A capture-ordered retrospective replay assesses task--model score-pattern consistency under scene renewal.

As shown in Figure \ref{fig:tasks}, using ten scenes from this inventory, we construct \textit{Start-10 full}, comprising 35,491 question records across 17 tasks spanning 2D perception, 3D perception, cross-view perception, and reasoning. A 928-question minival supports task examples and reasoning diagnostics. We evaluate 30 general-purpose, remote-sensing, and embodied models. All 18 tested embodied models outperform the three selected remote-sensing comparators in Overall; the strongest embodied model scores 43.29, compared with 59.88 for GPT-6-Astra. Fine-grained vehicle recognition remains challenging, and performance varies across observation heights and pitches.



Our contributions are threefold:
\begin{enumerate}
    \item \textbf{Fresh physical observations for continual aerial evaluation.} We introduce a live benchmark driven by recurring acquisition of real UAV imagery, with capture-time admission, scene-wise renewal, delayed answer disclosure, and versioned scoring.
    
    \item \textbf{Entity-centered spatial answer production.} We develop a geometry-assisted workflow that associates repeated observations with physical vehicles, reuses human-verified attributes, and preserves answer provenance for objectively scored spatial questions.
    
    \item \textbf{Systematic ASI evaluation across model families.} We evaluate 30 general-purpose, remote-sensing, and embodied models across 17 tasks, characterizing their profiles, fine-grained recognition difficulties, and performance under varying aerial observation conditions.
\end{enumerate}
\section{Related Work}
\label{sec:related_work}

\subsection{Live Benchmarks}

Live benchmarks refresh evaluation data to mitigate test-set contamination and maintain meaningful comparisons as models evolve. LiveBench and LiveCodeBench periodically introduce questions with objectively verifiable answers from recent information sources and programming contests, respectively~\cite{white2025livebench,jain2025livecodebench}. LiveXiv and MMBench-Live extend continual evaluation to scientific documents and multimodal understanding through recurring acquisition and verifiable question generation~\cite{shabtay2025livexiv,mmbench_live}. Extending this paradigm to ASI requires physically acquiring new observations and reconstructing and verifying geometric references and cross-view identities. RSLiveBench-UAV addresses these challenges through recurring UAV acquisition, geometry-assisted entity association, human-verified spatial references, and versioned evaluation.

\subsection{Aerial Spatial Intelligence}

ASI encompasses object semantics, physical properties, and spatial relationships from aerial viewpoints. General-purpose benchmarks such as BLINK, VSI-Bench, and MMSI-Bench evaluate visual perception, spatial reasoning, and multi-image understanding~\cite{fu2024blink,yang2025thinking,yang2026mmsibench}. Complementary work on embodied world models connects environmental prediction with action planning: DAWN jointly refines world and action predictions, while RISE adaptively allocates future rollout computation for autonomous driving~\cite{lu2026dawn,lu2026rise}. In aerial settings, SpatialSky-Bench, SpatialUAV, and Geo3DVQA evaluate UAV scene understanding, aerial spatial reasoning, and geospatial perception~\cite{zhang2025skyready,zhang2026spatialuav,tsujimoto2026geo3dvqa}. AVI-Math and AirSpatialBot further explore vehicle-centered mathematical reasoning and fine-grained spatial attributes in aerial imagery~\cite{zhou2025avimath,zhou2025airspatialbot}. These efforts advance spatial evaluation and embodied modeling. RSLiveBench-UAV extends this work with live ASI evaluation driven by continuously acquired UAV observations.

\begin{figure*}[!t]
\centering
\includegraphics[width=\textwidth]{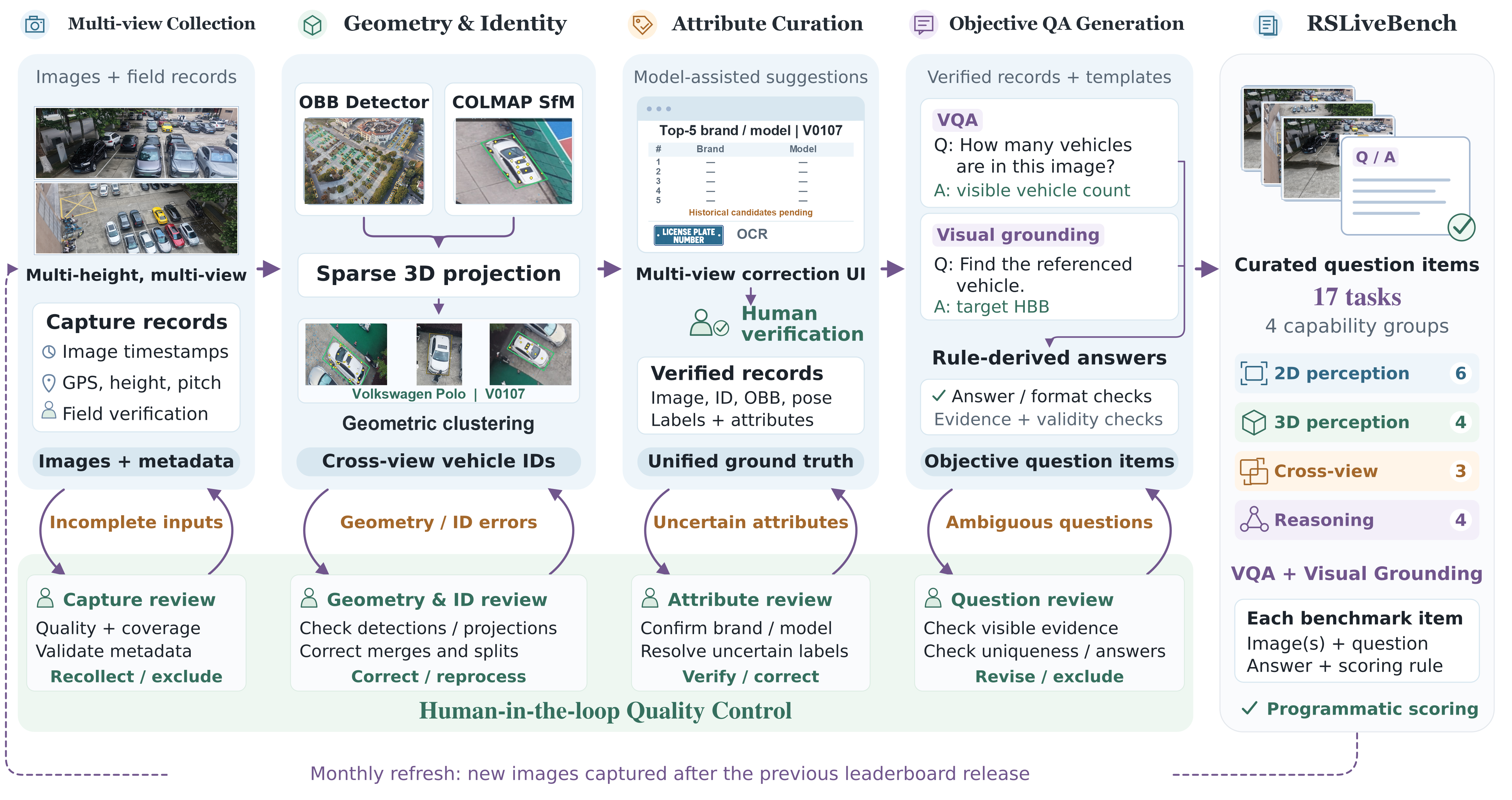}
\caption{The Entity-centered answer-production workflow.}
\label{fig:pipeline}
\end{figure*}

\subsection{Remote-Sensing Vision-Language Benchmarks}

Remote-sensing vision-language benchmarks have expanded in task diversity, grounding, data quality, and evaluation coverage~\cite{zhou2026datadriven}. RSVQA and RSVG establish visual question answering and referring-expression grounding~\cite{lobry2020rsvqa,zhan2023rsvg}, while VRSBench and XLRS-Bench broaden coverage to comprehensive image understanding, grounding, and extremely high-resolution imagery~\cite{li2024vrsbench,wang2025xlrsbench}. CHOICE provides hierarchical capability evaluation, whereas ChatEarthBench emphasizes independently collected Earth-observation imagery~\cite{an2025choice,yuan2026chatearthbench}. Many remote-sensing benchmarks use static test sets. RSLiveBench-UAV complements them with newly captured aerial imagery and delayed answer disclosure, supporting recurring evaluation while mitigating contamination risk.
\section{RSLiveBench-UAV}
\label{sec:method}

RSLiveBench-UAV connects scene renewal, entity-centered ground-truth production, and model evaluation. Scenes define the update unit, physical vehicle instances organize reusable attributes, and questions provide the inputs and references for evaluation.

\subsection{Acquisition-Constrained Live Protocol}
\label{sec:live_protocol}

To tie benchmark renewal to fresh UAV observations, we define a scene as a timestamped field-acquisition batch. Building on continual evaluation~\cite{white2025livebench}, we maintain a four-scene first-in, first-out (FIFO) window. Let $v$ index benchmark versions, with $R_v$ denoting the release time of version $v$ and $\mathcal{W}_v$ its active scene roster, where $|\mathcal{W}_v|=4$. After initialization, each update admits a newly acquired scene $a_v$ and retires the earliest-admitted active scene $r_v$:
\begin{equation}
\begin{gathered}
R_{v-1}<\tau(I)<R_v,\qquad I\in\mathcal{I}(a_v),\\
\mathcal{W}_v=(\mathcal{W}_{v-1}\setminus\{r_v\})\cup\{a_v\}.
\end{gathered}
\label{eq:live_admission_window}
\end{equation}
Here, $\tau(I)$ is the capture time of image $I$, and $\mathcal{I}(a_v)$ is the incoming image set. Each monthly update retains three scenes and replaces one; four consecutive replacements retire the entire initial roster. This 25\% renewal rate applies to scenes, while question counts and within-component score contributions may vary by scene.

Before release $R_v$, questions and references must be reviewed, and every declared task component must be covered. Each release records its scene roster, capture dates, component counts, answer/scorer versions, and model runs, allowing results to be traced to their scene composition and scoring configuration.

Full answers for a scene $s$ remain confidential until retirement and expiry of the minimum embargo from first admission:
\begin{equation}
\begin{aligned}
T_{\mathrm{public}}(s)\ge
\max\bigl\{&T_{\mathrm{retire}}(s),\\
&T_{\mathrm{first}}(s)+\Delta_{\mathrm{emb}}\bigr\}.
\end{aligned}
\label{eq:answer_disclosure}
\end{equation}
Here, $T_{\mathrm{first}}(s)$, $T_{\mathrm{retire}}(s)$, and $T_{\mathrm{public}}(s)$ denote first admission, retirement, and full-answer disclosure, respectively. The minimum embargo $\Delta_{\mathrm{emb}}$ is four calendar months.

Across the 102 collected scenes, acquiring UAV imagery took an average of 14.48 minutes per scene, with the acquisition-time distribution reported in Appendix~\ref{app:capture_time}. These records provide an empirical estimate of the acquisition effort required to add a new scene and support the feasibility of recurring field acquisition for monthly benchmark renewal.

\begin{table}[t]
\centering
\caption{Scene and image coverage with recorded annotation time by collection period. Hours are summed within each reporting period.}
\label{tab:production_cost}
\begingroup
\scriptsize
\setlength{\tabcolsep}{2pt}
\renewcommand{\arraystretch}{1.15}
\begin{tabular*}{\columnwidth}{@{\extracolsep{\fill}}lrrrrrr@{}}
\toprule
Year & Month & Scenes & Images & Total (h) & min/scene & s/image \\
\midrule
2023 & 9 & 11 & 623   & 24.99 & 136.31 & 144.40 \\
2026 & 7 & 7  & 1,195 & 5.17  & 44.33  & 15.58  \\
2026 & 8 & 26 & 3,529 & 15.08 & 34.80  & 15.38  \\
2026 & 9 & 52 & 6,815 & 25.58 & 29.52  & 13.51  \\
\bottomrule
\end{tabular*}
\endgroup
\end{table}

\subsection{Geometry-Assisted Ground-Truth Production}
\label{sec:production}

Continual benchmark renewal requires efficient ground-truth production. Historical image-by-image annotation averaged 136.31 minutes per scene, whereas the mean recorded with our entity-centered workflow was 29.52 minutes per active scene--month in September 2026, 78.3\% below the historical mean (Table~\ref{tab:production_cost}). The workflow reuses verified entity attributes across views while preserving observation-specific references and answer provenance. Figure~\ref{fig:pipeline} summarizes collection, association, attribute curation, and question generation.

\paragraph{Spatial mapping.}
Each batch contains multi-height, multi-view images with timestamps,
GPS coordinates, height/pitch metadata, and field-verification
records. Oriented bounding box (OBB)
detection~\cite{zhou2022mmrotate} and COLMAP-based
structure-from-motion (SfM) provide vehicle observations and sparse
scene geometry. We align the reconstruction to a metric
east--north--up (ENU) frame through a similarity transform based on
XMP geolocation and relative altitude. Anchors use nearby covisible
sparse points, falling back to ground projection when support is
insufficient. For detection $k$ in image $i$, let
$\tilde{\mathbf{u}}_{ik}=(u_{ik},v_{ik},1)^\top$
be its undistorted homogeneous OBB center. With camera intrinsics
$\mathbf{K}_i$, world-to-camera rotation $\mathbf{R}_i$, and camera
center $\mathbf{C}_i$ in the aligned frame, its fallback ground
projection is
\begin{equation}
\mathbf{q}_{ik}=\mathbf{C}_i+\lambda_{ik}\mathbf{R}_i^\top
\mathbf{K}_i^{-1}\tilde{\mathbf{u}}_{ik}.
\label{eq:spatial_mapping}
\end{equation}
The positive depth $\lambda_{ik}$ is selected so that
$\mathbf{q}_{ik}\in\mathcal{G}$, where $\mathcal{G}$ is the batch's
ground reference in the aligned frame. These approximate spatial
anchors supply geometric evidence for cross-view association,
which is subsequently checked against image observations.

\paragraph{Constrained spatial clustering.}
We group mapped detections greedily in descending confidence,
assigning each to the nearest eligible cluster centroid within
a scene-adaptive XY distance threshold $\epsilon$, or creating
a new cluster if none exists. Cluster membership obeys a
same-image cannot-link constraint:
\begin{equation}
\begin{aligned}
d(a,e)&=\|\mathbf{q}_a^{\mathrm{xy}}-\boldsymbol{\mu}_e\|_2,\\
\mathcal{C}_e&=\{a:z_a=e\},\quad
|\mathcal{C}_e\cap\mathcal{D}_i|\leq 1.
\end{aligned}
\label{eq:spatial_clustering}
\end{equation}
Here, $a=(i,k)$ indexes a detection with XY anchor
$\mathbf{q}_a^{\mathrm{xy}}$ and candidate label $z_a$,
$\boldsymbol{\mu}_e$ is the current XY centroid of cluster $e$,
and $\mathcal{D}_i$ contains detections from image $i$.
Geometry-based postprocessing further merges nearby clusters
using predicted merge probabilities while preserving this
constraint. Reviewers correct detections, projections, and
identity merges or splits before accepting associations.
Once verified, each cluster defines a physical entity whose
member images supply observations for attribute curation and
question generation. For metric spatial references, a separate
iTwin Capture Modeler branch uses photograph GPS and UAV poses
to reconstruct georeferenced meshes and refine camera poses.

\paragraph{Attribute verification and reuse.}
Associated views support Top-5 brand/model suggestions, with optical character recognition (OCR) as auxiliary evidence. Reviewers verify identities and attributes using crops, original images, projections, and source records. Vehicle dimensions are specification-based attributes retrieved by confirmed brand/model from Dongchedi\footnote{\url{https://www.dongchedi.com/}} and Autohome\footnote{\url{https://www.autohome.com.cn/car/}}. Within each verified batch, entity $e$ has shared attributes $\mathbf{a}_e$ and verified visible views $\mathcal{V}_e$:
\begin{equation}
\begin{aligned}
\mathcal{E}_e&=\left(\mathbf{a}_e,\{o_{i,e}\}_{i\in\mathcal{V}_e}\right),\\
o_{i,e}&=\left(i,B_{i,e},\nu_{i,e},\mathbf{c}_i\right).
\end{aligned}
\label{eq:entity_observation_records}
\end{equation}
Here, $\mathbf{a}_e$ stores the confirmed brand/model and specifications. Each observation $o_{i,e}$ stores image index $i$, horizontal bounding box (HBB) $B_{i,e}$, visibility $\nu_{i,e}$, and camera metadata $\mathbf{c}_i$. This separation supports attribute reuse while keeping visibility and image-plane grounding specific to each observation. Reuse is restricted to verified within-batch identities; new batches require renewed identity verification.

\paragraph{Question generation and provenance.}
Templates combine verified entities, observations, and spatial references to construct fixed-format questions and answers. Shared identities determine which vehicles are present across views, while image-specific observations supply their grounding references. Let $\mathcal{S}_i=\{e:i\in\mathcal{V}_e\}$ denote the verified visible entities in image $i$. Cross-view union/intersection counts and instance re-identification (RI) references are
\begin{equation}
\begin{aligned}
n^{\cup}_{ij}&=|\mathcal{S}_i\cup\mathcal{S}_j|,\quad
n^{\cap}_{ij}=|\mathcal{S}_i\cap\mathcal{S}_j|,\\
y^{\mathrm{RI}}_{i\rightarrow j}(e)&=B_{j,e},\quad
e\in\mathcal{S}_i\cap\mathcal{S}_j.
\end{aligned}
\label{eq:entity_grounded_answers}
\end{equation}
The two counts measure distinct vehicles visible in either image and shared vehicles visible in both, respectively. RI returns the target-image HBB of the specified source entity, verified as visible in both images. Single-image counting returns $|\mathcal{S}_i|$. Cross-view grounding (CG) requires a unique common entity, $\mathcal{S}_i\cap\mathcal{S}_j=\{e^\star\}$, and returns its target-image HBB $B_{j,e^\star}$. All grounding references use HBBs for a common output format across models.

Each question retains supporting observations, entity identities, reference sources, and scorer configurations. Human review checks target identity, uniqueness, visual evidence, reference consistency, and format; ambiguous items are revised or excluded.

\subsection{Spatial Tasks and Deterministic Evaluation}
\label{sec:tasks_scoring}

\paragraph{Spatial tasks.}
The resulting questions support evaluation of ASI across 17 tasks in four complementary capability groups (Fig.~\ref{fig:tasks}). \textbf{2D perception} assesses semantic and image-plane understanding; \textbf{3D perception} examines physical dimensions and metric spatial relationships; \textbf{cross-view perception} evaluates identity association and set-based reasoning across views; and \textbf{reasoning} assesses viewpoint geometry and capture context. Across these groups, three tasks use categorical visual question answering (VQA), ten use numerical VQA, and four use visual grounding. BI and BV receive vehicle crops, while other tasks receive original images; embedded metadata are removed from all input images. BI uses view-level scoring units, BV uses vehicles with complete expected-view sets, and other tasks use individual questions.

\begin{table*}[!t]
\centering
\caption{Scores for 12 representative models on Start-10 full. Overall equally weights the four capability-group means. Bold and underlining mark the highest and second-highest values among displayed models.}
\label{tab:model_overview}
\begingroup
\small
\setlength{\tabcolsep}{7pt}
\renewcommand{\arraystretch}{1.02}
\begin{tabular*}{\textwidth}{@{\extracolsep{\fill}}lrrrrr@{}}
\toprule
Model & Overall & 2D & 3D & Cross-view & Reasoning \\
\midrule
\multicolumn{6}{l}{\textbf{Remote sensing MLLMs}} \\
VHM~\cite{pang2025vhm} & 4.38 & 9.83 & 0.33 & 6.48 & 0.89 \\
GeoChat-7B~\cite{kuckreja2024geochat} & 8.31 & 8.63 & 7.15 & 7.64 & 9.82 \\
\midrule
\multicolumn{6}{l}{\textbf{Embodied MLLMs}} \\
SenseNova-SI-1.4-InternVL3-8B~\cite{cai2026sensenovasi} & 18.95 & 22.16 & 12.01 & 23.36 & 18.30 \\
Cosmos-Reason2-8B~\cite{nvidia2025cosmosreason2} & 20.80 & 27.02 & 17.42 & 13.71 & \underline{25.05} \\
MiMo-Embodied-7B~\cite{hao2025mimoembodied} & 23.98 & 23.68 & 42.07 & 11.87 & 18.30 \\
RoboBrain2.5-8B-NV~\cite{tan2026robobrain25} & 32.84 & 35.55 & 57.49 & 16.09 & 22.24 \\
RynnBrain1.1-9B~\cite{li2026rynnbrain11} & 35.10 & 31.13 & 56.73 & 37.79 & 14.75 \\
ACE-Brain-0.6-9B~\cite{wang2026acebrain06} & \underline{43.29} & 33.96 & \underline{69.52} & \underline{55.97} & 13.70 \\
\midrule
\multicolumn{6}{l}{\textbf{General-purpose MLLMs}} \\
InternVL3.5-4B~\cite{wang2025internvl35} & 18.81 & 19.97 & 27.73 & 11.56 & 15.98 \\
Qwen3.5-9B~\cite{qwen2026qwen35} & 36.08 & \underline{40.20} & 69.31 & 15.69 & 19.10 \\
DeepSeek-V4.1-Flash-1017-low~\cite{deepseek2026v41flash1017} & 37.19 & 36.34 & 60.70 & 32.38 & 19.33 \\
GPT-6-Astra-low~\cite{openai2026gpt} & \textbf{59.88} & \textbf{54.17} & \textbf{77.46} & \textbf{69.07} & \textbf{38.81} \\
\bottomrule
\end{tabular*}
\endgroup
\end{table*}

\paragraph{Deterministic evaluation.}
Task definitions, output formats, and scoring rules remain fixed across releases. Deterministic scorers assign scores from 0 to 100 without model-based judges. For applicable numerical components with positive references, we use mean relative accuracy (MRA) from VSI-Bench~\cite{yang2025thinking}, averaging accuracy over relative-error thresholds from 5\% to 50\% in 5\% increments. Visual grounding uses HBB accuracy at an intersection-over-union (IoU) threshold of $0.5$. Angular and geographic components average accuracies at $5^\circ$, $10^\circ$, and $15^\circ$, and at 1, 10, and 100\,km, respectively. To obtain component scores, we pool scoring units across scenes and average their scores. Equal averaging then proceeds over components within each task, tasks within each capability group, and the four groups to obtain Overall.

\paragraph{Consistency metrics.}
Following MMBench-Live~\cite{mmbench_live}, we use Pearson correlation $r$ and Spearman rank correlation $\rho$ on task-level scores to assess consistency across scene collections and adjacent benchmark versions. For each comparison, we align the $N$ task--model entries shared by the two evaluations to form score vectors $\mathbf{x}$ and $\mathbf{y}$. For $N\geq2$ and nonconstant score vectors, the metrics are
\begin{equation}
\begin{aligned}
r(\mathbf{x},\mathbf{y})
&=\frac{\sum_{\ell=1}^{N}(x_\ell-\bar{x})(y_\ell-\bar{y})}
{\sqrt{\sum_{\ell=1}^{N}(x_\ell-\bar{x})^2}
\sqrt{\sum_{\ell=1}^{N}(y_\ell-\bar{y})^2}},\\
\rho(\mathbf{x},\mathbf{y})
&=r\!\left(\operatorname{rank}(\mathbf{x}),\operatorname{rank}(\mathbf{y})\right).
\end{aligned}
\label{eq:consistency_metrics}
\end{equation}
Here, $N$ is the number of shared task--model entries, $\ell$ indexes a matched entry, and $x_\ell$ and $y_\ell$ are its scores in the two evaluations. The means of the respective score vectors are $\bar{x}$ and $\bar{y}$. The operator $\operatorname{rank}$ replaces each score by its rank within the vector, assigning average ranks to ties while preserving entry alignment. Pearson measures linear association between score profiles, while Spearman directly quantifies rank consistency of the task--model entries. Section~\ref{sec:stability} reports both metrics under scene replacement and incremental renewal.

\section{Experiments}
\label{sec:experiments}

\subsection{Evaluation Setup}
\label{sec:experimental_setup}
\paragraph{Data and coverage.}
Main results use Start-10 full, comprising 35,491 question records from ten scenes and covering 17 tasks. The 928-question minival of Start-10 supports reasoning-effort diagnostics. Start-10 is a ten-scene evaluation subset of the 102-scene acquisition inventory, whereas History-11 contains separately collected historical scenes. Prospective releases are outside the present experiments.

\paragraph{Models and inputs.}

We evaluate three remote-sensing, 18 embodied, and nine general-purpose Multimodal Large Language Models (MLLMs). The three remote-sensing MLLMs are GeoLLaVA-8K~\cite{wang2025geollava8k}, VHM~\cite{pang2025vhm}, and GeoChat-7B~\cite{kuckreja2024geochat}.
The 18 embodied brains are SenseNova-SI-1.1-InternVL3-2B/8B and SenseNova-SI-1.3/1.4/1.5-InternVL3-8B~\cite{cai2026sensenovasi}; Pelican1.0-VL-7B (2K); RynnBrain-2B/8B and RynnBrain1.1-2B/9B~\cite{li2026rynnbrain11}; Cosmos-Reason2-2B/8B~\cite{nvidia2025cosmosreason2}; ACE-Brain-0-8B~\cite{gong2026acebrain0}, ACE-Brain-0.5-8B~\cite{ace2026brain05}, and ACE-Brain-0.6-9B~\cite{wang2026acebrain06}; MiMo-Embodied-7B~\cite{hao2025mimoembodied}; PhysBrain1.5-8B~\cite{deepcybo2026physbrain15}; and RoboBrain2.5-8B-NV~\cite{tan2026robobrain25}.
The nine general-purpose MLLMs are Qwen3.5-2B/9B~\cite{qwen2026qwen35}, InternVL3.5-4B/8B~\cite{wang2025internvl35}, Qwen3-VL-8B-Instruct~\cite{bai2025qwen3vl}, DeepSeek-V4.1-Flash-1017-low~\cite{deepseek2026v41flash1017}, and GPT~\cite{openai2026gpt}. Models receive question text and image(s), with embedded metadata removed. BI/BV use vehicle crops; other tasks use original images without evaluator-added visual markers. Multi-image inputs are supplied separately and simultaneously. Original capture metadata are retained only for reference construction and diagnostic grouping.

\subsection{Main Results}
\label{sec:model_results}

Table~\ref{tab:model_overview} reports Overall and four capability-group scores for 12 representative models, while complete results for all 30 models appear in Appendix~\ref{app:model_results}.

\paragraph{Advanced general-purpose models.}
Among the displayed representatives, GPT-6-Astra leads all four capability groups, exceeding the next-highest representative by 16.59 Overall points. Qwen3.5-9B and DeepSeek-V4.1-Flash-1017 have similar Overall scores but distinct profiles: Qwen leads in 3D, while DeepSeek leads in Cross-view.

\paragraph{Embodied models.}
ACE-Brain-0.6-9B leads the six embodied representatives with 43.29 Overall, exceeding Qwen3.5-9B by 7.21 points. The two models achieve nearly identical 3D scores (69.52 versus 69.31), while ACE scores 40.28 points higher in Cross-view. Because ACE scores lower in both 2D and Reasoning, its Overall advantage is driven primarily by cross-view perception. The ranking changes markedly in Reasoning: Cosmos-Reason2-8B leads the embodied representatives with 25.05, exceeding both Qwen and DeepSeek despite its substantially lower Overall score of 20.80, whereas ACE ranks last among the six with 13.70. This reversal makes viewpoint and capture-context reasoning an essential consideration when assessing embodied models for aerial spatial understanding.


%

\paragraph{Remote-sensing models.}
GeoChat-7B and VHM obtain Overall scores of 8.31 and 4.38, respectively, ranking below all other displayed models. Despite their remote-sensing specialization---with GeoChat's instruction-tuning data, for example, including examples drawn from aerial-image datasets---both models score below 10 in all four capability groups. These results suggest that exposure to overhead imagery alone does not ensure strong performance on the object-level aerial spatial tasks evaluated here.
%



\subsection{Reliability of Rolling Evaluation}
\label{sec:stability}

\paragraph{Window size selection.}
We first examine how many scenes a benchmark version should contain to obtain consistent task--model performance patterns. For each window size $k$, we compare all unordered pairs of disjoint scene subsets using shared model--task scores from all 30 models. Agreement increases as more scenes are aggregated, with smaller additional gains beyond four scenes (Table~\ref{tab:window_sizes}). This supports our four-scene window as a practical balance between cross-version consistency and data renewal. The reported standard deviations summarize variation across subset pairs, which reuse scenes.

\begin{table}[t]
\centering
\caption{Task--model score consistency across all unordered pairs of disjoint $k$-scene subsets from the RSLiveBench Start-10 scenes across pairs for all 30 models.}
\label{tab:window_sizes}
\small
\setlength{\tabcolsep}{3pt}
\renewcommand{\arraystretch}{1.12}
\begin{tabular*}{\columnwidth}{@{\extracolsep{\fill}}cccc@{}}
\toprule
\multicolumn{2}{c}{Comparison setting} & \multicolumn{2}{c}{Correlation} \\
\cmidrule(lr){1-2}\cmidrule(l){3-4}
Grouping & Pairs & Pearson $r$ & Spearman $\rho$ \\
\midrule
$k=1$ & 45 & \makecell[c]{0.7449\\{\footnotesize\textcolor{cvprblue}{$\pm 0.1072$}}} & \makecell[c]{0.7369\\{\footnotesize\textcolor{cvprblue}{$\pm 0.0928$}}} \\
$k=2$ & 630 & \makecell[c]{0.8475\\{\footnotesize\textcolor{cvprblue}{$\pm 0.0605$}}} & \makecell[c]{0.8263\\{\footnotesize\textcolor{cvprblue}{$\pm 0.0589$}}} \\
$k=3$ & 2,100 & \makecell[c]{0.8962\\{\footnotesize\textcolor{cvprblue}{$\pm 0.0450$}}} & \makecell[c]{0.8818\\{\footnotesize\textcolor{cvprblue}{$\pm 0.0406$}}} \\
$k=4$ & 1,575 & \makecell[c]{0.9233\\{\footnotesize\textcolor{cvprblue}{$\pm 0.0355$}}} & \makecell[c]{0.9144\\{\footnotesize\textcolor{cvprblue}{$\pm 0.0298$}}} \\
$k=5$ & 126 & \makecell[c]{0.9393\\{\footnotesize\textcolor{cvprblue}{$\pm 0.0289$}}} & \makecell[c]{0.9345\\{\footnotesize\textcolor{cvprblue}{$\pm 0.0235$}}} \\
\midrule
MMBench-Live~\cite{mmbench_live} & 2 & 0.5200 & 0.4254 \\
\bottomrule
\end{tabular*}
\end{table}

\begin{figure}[tbp]
\centering
\includegraphics[width=\columnwidth]{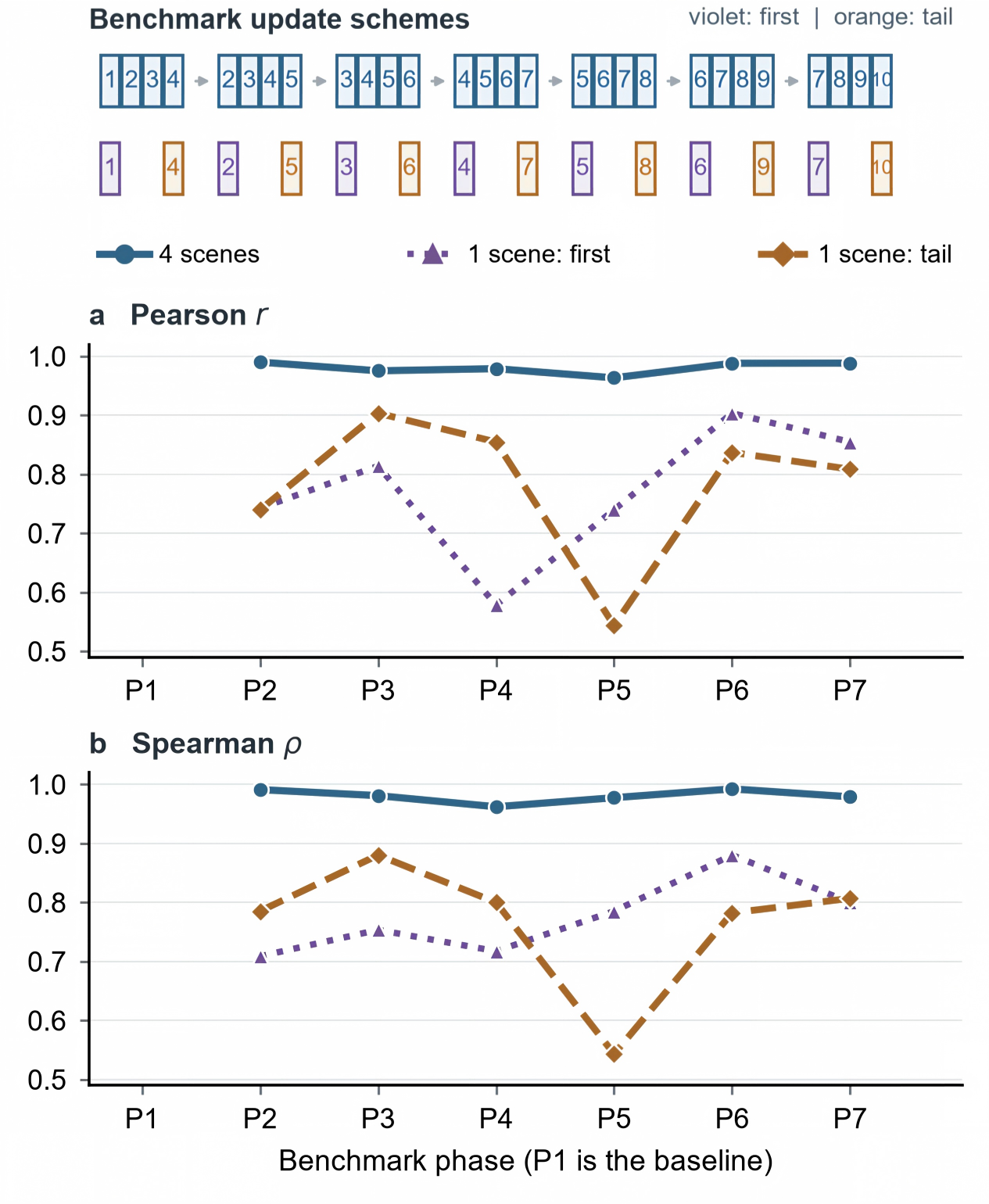}
\caption{Adjacent task--model agreement in a seven-release retrospective replay. Four-scene neighbors share three scenes; single-scene controls compare their first or tail scenes.}
\label{fig:adjacent_consistency}
\end{figure}


\paragraph{Consistency under rolling updates.}
We next test how the selected window behaves when scenes are progressively replaced. A retrospective replay through capture-ordered Start-10 yields seven four-scene versions and six adjacent transitions. Spearman correlation assesses the relative ordering of matched task–model entries. Both metrics remain higher and less variable than those of the first- and tail-scene controls (Figure~\ref{fig:adjacent_consistency}), indicating greater continuity of the pooled task–model score profiles. Retaining three scenes while introducing one new scene smooths changes in the measured performance profile. Together, the experiments assess complementary properties: agreement across non-overlapping scene collections and continuity during incremental renewal. These correlations quantify task--model score-pattern agreement; the History-11 comparison is reported separately in Appendix~\ref{app:year_rankings}.

\subsection{Diagnostic Analysis}
\label{sec:ablation_study}
\label{sec:geometry_diagnostics}

\begin{figure}[t]
\centering
\includegraphics[width=\columnwidth]{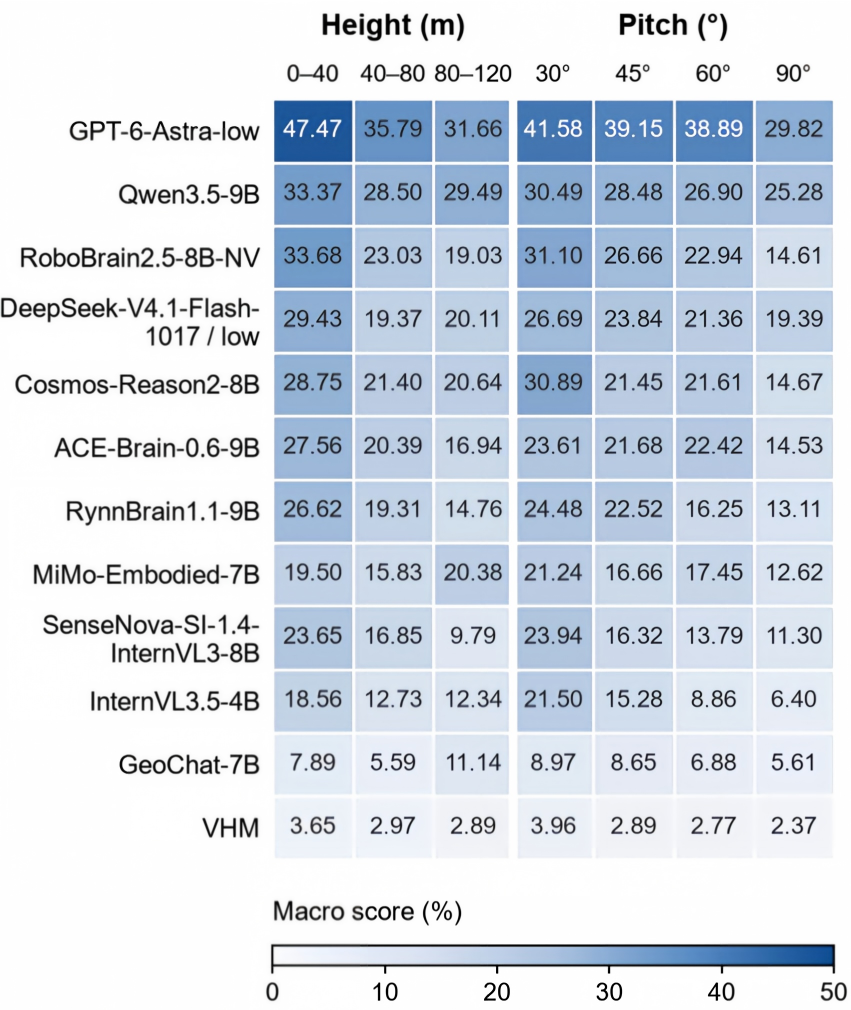}
\caption{Task-macro-average scores by height (left) and pitch (right) for 12 representative models.}
\label{fig:geometry}
\end{figure}

\paragraph{Sensitivity to aerial observation conditions.}
Figure~\ref{fig:geometry} reports an equal-weight average over eight diagnostic tasks (BI, BV, WU, RD, CP, PC, CL, and CH). Ten of twelve representatives score lower in the highest than the lowest height bin; all twelve decline from $30^\circ$ to $90^\circ$. Astra also loses 15.81 and 11.76 points across the respective endpoints, showing that observation geometry affects even the strongest models. Consistent with expectations, lower flight heights and smaller downward pitch angles yield views closer to conventional ground-level perspectives and are generally associated with higher model scores. RSLiveBench-UAV captures this condition-dependent performance through multi-height, multi-pitch sampling.

\label{sec:bi_bv_comparison}
\begin{figure}[tbp]
\centering
\includegraphics[width=\columnwidth]{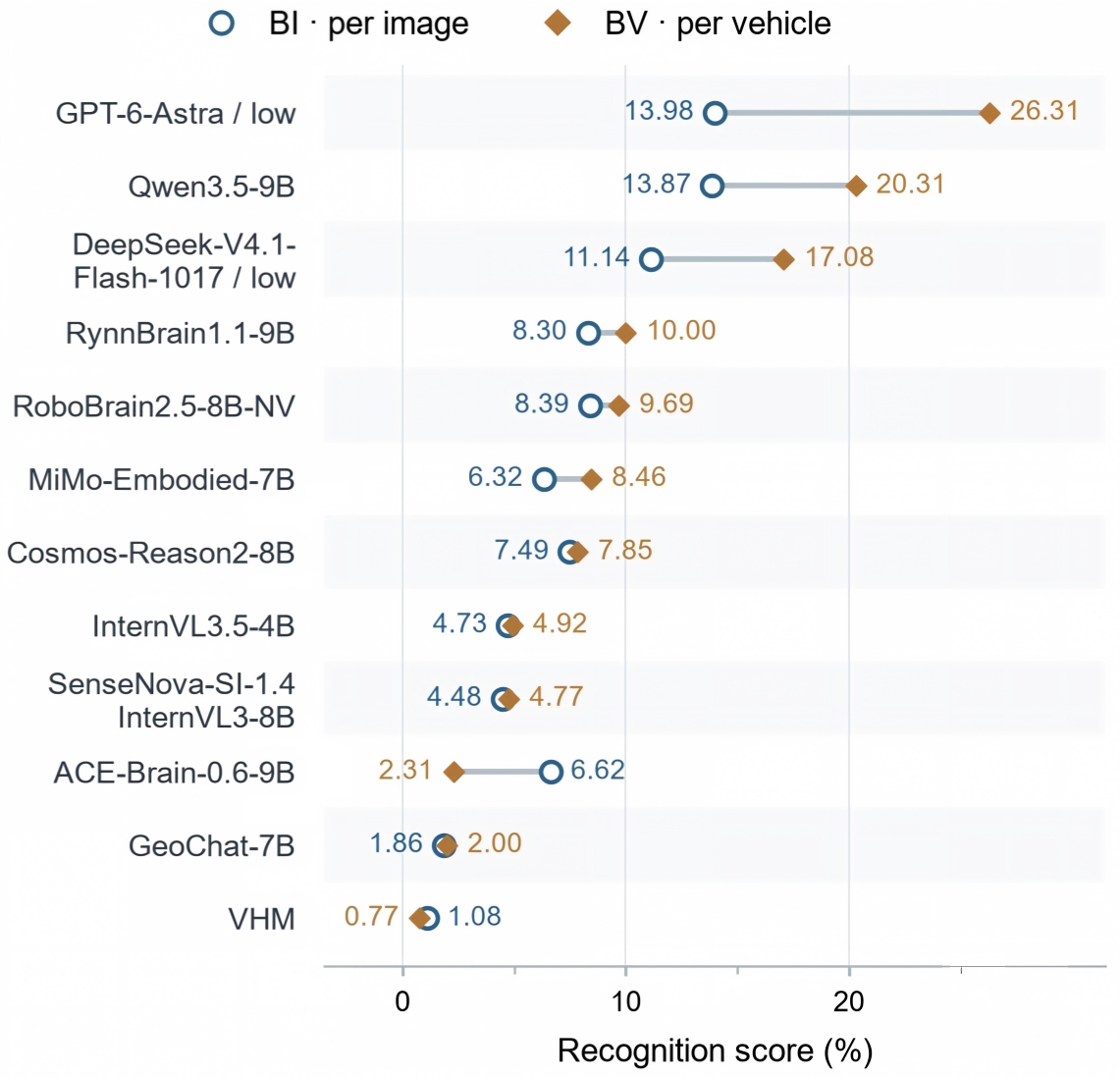}
\caption{Per-image (BI) and per-vehicle (BV) recognition for 12 representatives. Both combine brand and joint brand--model accuracy; BV votes over independent crop predictions.}
\label{fig:bi_bv_representative12}
\end{figure}

\paragraph{Vehicle model recognition from aerial views.}
Across all 30 models, BV exceeds BI for 21; the highest BV score remains below 27 under the combined brand and joint brand--model metric. Astra and Qwen3.5-9B nearly tie in BI, while Astra leads BV by 6 points (Figure~\ref{fig:bi_bv_representative12}). Equal-vehicle averaging preserves the BI near-tie; voting adds 12.33 points for Astra and 6.44 for Qwen3.5, indicating that Astra benefits more from aggregating independent predictions. Absolute scores remain low, leaving substantial headroom in fine-grained aerial vehicle recognition.

\begin{figure}[tbp]
\centering
\includegraphics[width=\columnwidth]{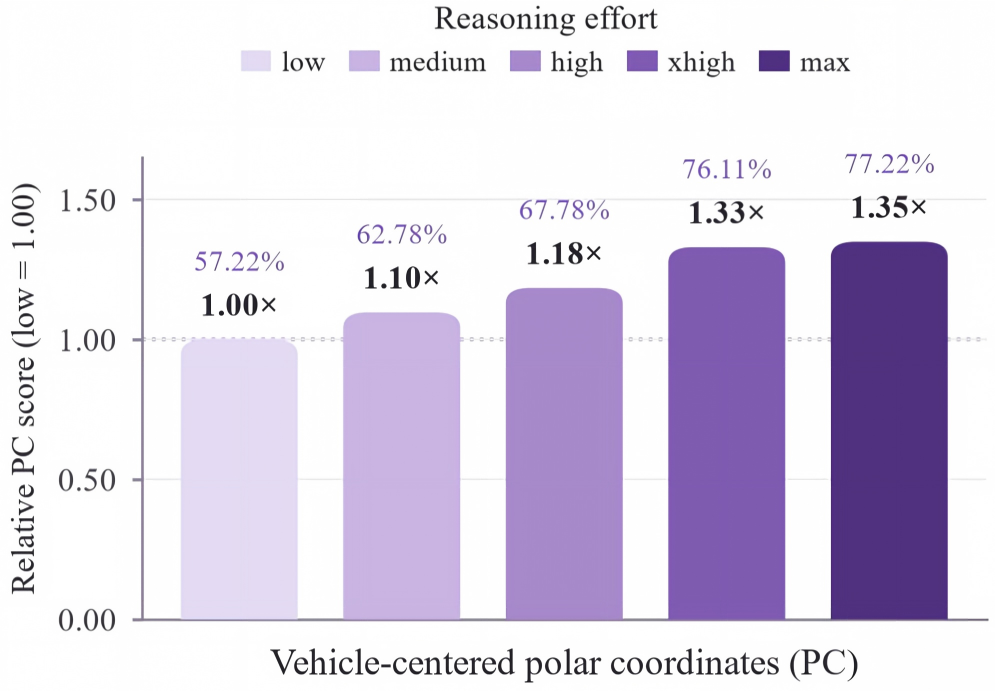}
\caption{GPT-6-Astra's minival PC score across five reasoning-effort settings, with tools disabled.}
\label{fig:astra_pc_reasoning_effort}
\end{figure}

\paragraph{Spatial imagination and reasoning effort.}
PC probes spatial imagination by requiring a shift from the observed aerial view to a vehicle-centered reference frame. Taking the target vehicle as the origin, the model must infer the UAV's azimuth and elevation from visual cues. This makes PC a challenging test of three-dimensional spatial reasoning beyond recognizing objects or locating them in the image plane. On minival with tools disabled, Astra's score increases monotonically from 57.22 at low to 77.22 at max across five reasoning-effort settings (Figure~\ref{fig:astra_pc_reasoning_effort}). Higher reasoning effort improves PC performance, although the gain narrows to 1.11 points from xhigh to max.
\section{Limitations and Discussion}
\label{sec:limitations}

RSLiveBench-UAV evaluates individual spatial skills rather than compound tasks requiring coordinated capabilities. Its offline, single-turn image interface excludes video understanding, continuous spatial-state updates, and persistent tracking. The vehicle focus also limits object diversity, although model specifications provide low-cost dimension references that support recurring data production. Nevertheless, the benchmark spans four complementary capability groups. Focused questions support capability-level diagnosis, while the image-based interface enables reuse of existing evaluation pipelines with limited adaptation. This provides an accessible foundation for aerial evaluation of embodied models, rather than a complete assessment of autonomous aerial intelligence. Future extensions could diversify object categories and introduce agent-based evaluation through compound tasks, sequential observations, and multi-step interaction. These extensions would examine how agents combine spatial skills, actively acquire observations, and update their understanding to guide decisions.
\section{Conclusion}
\label{sec:conclusion}
RSLiveBench-UAV links fresh UAV observations, entity-centered answer production, and deterministic evaluation within a versioned live protocol. Across 30 tested systems, embodied models exceed the selected remote-sensing comparators in Overall; recognition and spatial understanding remain key areas for improvement. Retrospective scene-group and rolling-window analyses characterize score-pattern agreement under scene replacement. Together, the protocol, production workflow, and evaluation establish a practical foundation for recurring aerial assessment. Prospective deployment and independent quality audits are the next steps for assessing long-term reliability.

{
    \small
    \bibliographystyle{ieeenat_fullname}
    \bibliography{main}
}

\clearpage
\setcounter{page}{1}
\renewcommand{\thepage}{S\arabic{page}}
\setcounter{equation}{0}
\setcounter{figure}{0}
\setcounter{table}{0}
\renewcommand{\theequation}{S\arabic{equation}}
\renewcommand{\thefigure}{S\arabic{figure}}
\renewcommand{\thetable}{S\arabic{table}}
\renewcommand{\theHequation}{supp.\arabic{equation}}
\renewcommand{\theHfigure}{supp.\arabic{figure}}
\renewcommand{\theHtable}{supp.\arabic{table}}
\raggedbottom
\maketitlesupplementary
\label{start:supplement}
\appendix

\section{Complete Scoring and Aggregation Rules}
\label{app:scoring}

This appendix specifies the deterministic scoring protocol for all 17 tasks. Scoring functions, answer types, legal ranges, and periodicity settings are fixed by question configurations rather than inferred from model responses. Unless noted otherwise, primary scores range from 0 to 100.

\subsection{Scoring Units and Task Components}
\label{app:score_components}

A component groups observations scored by the same rule. BI uses vehicle views as units; BV uses physical vehicles with complete expected-view sets; other tasks use individual questions. Components within each task are equally weighted. BI and BV use brand and joint brand--model accuracy; SC uses count; SR uses grid and extreme positions; WU uses category; VH uses heading; LW uses length and width; HT uses height; RD uses nearest and farthest targets; AD uses distance; CC uses union, intersection, and difference; CG and RI use target grounding; CP uses height and pitch; PC uses azimuth and elevation; CL uses GPS location; and CH uses capture hour.

\begin{table}[tb]
\centering
\caption{Task components and primary scoring families. Numerical rules follow the frozen component configuration; MRA is defined for positive references.}
\label{tab:score_components}
\footnotesize
\setlength{\tabcolsep}{3pt}
\begin{tabular}{@{}lp{0.44\columnwidth}p{0.33\columnwidth}@{}}
\toprule
Task & Components & Scoring family\\
\midrule
BI & Brand; brand--model pair & Exact label matches\\
BV & Voted brand; voted pair & Exact label matches\\
SC & Count & Configured count score\\
SR & Grid; extreme position & HBB Acc@0.5\\
WU & Weather category & Exact category\\
VH & Heading & Angle thresholds\\
LW & Length; width & MRA\\
HT & Height & MRA\\
RD & Nearest; farthest & HBB Acc@0.5\\
AD & Camera--anchor range & Configured numerical score\\
CC & Union; intersection; difference & Configured count score\\
CG & Unique common target & HBB Acc@0.5\\
RI & Specified source target & HBB Acc@0.5\\
CP & Camera height; pitch & Numerical; angle\\
PC & Azimuth; elevation & Angle thresholds\\
CL & Latitude/longitude & Geographic thresholds\\
CH & Capture hour & Configured exact match\\
\bottomrule
\end{tabular}
\end{table}

\subsection{Brand and Model Recognition}
\label{app:recognition_scores}

For view $i$, reference and predicted labels each consist of a brand and a model. Before comparison, labels undergo Unicode case folding, leading and trailing whitespace removal, and normalization of consecutive whitespace to one space. No alias substitution or fuzzy matching is applied to the reported resultse. Brand accuracy and joint brand--model accuracy are
\begin{equation}
\begin{aligned}
A_{\mathrm{brand}}&=\frac{100}{N}\sum_{i=1}^{N}\mathbb{1}[\hat b_i=b_i],\\
A_{\mathrm{pair}}&=\frac{100}{N}\sum_{i=1}^{N}
 \mathbb{1}[(\hat b_i,\hat m_i)=(b_i,m_i)],\\
T_{\mathrm{BI}}&=\frac{A_{\mathrm{brand}}+A_{\mathrm{pair}}}{2}.
\end{aligned}
\label{eq:brand_image}
\end{equation}

A correct brand with an incorrect model contributes 50 points to the two-component mean. BV first aggregates predictions over all expected views of a vehicle, voting separately for the brand and the brand--model pair. Ties are broken by choosing the lexicographically smallest normalized label, and the tie is recorded:
\begin{equation}
\begin{aligned}
A_{\mathrm{brand}}^{\mathrm{veh}}&=\frac{100}{|V|}\sum_{v\in V}\mathbb{1}[\bar b_v=b_v],\\
A_{\mathrm{pair}}^{\mathrm{veh}}&=\frac{100}{|V|}\sum_{v\in V}\mathbb{1}[\bar q_v=q_v],\\
T_{\mathrm{BV}}&=\frac{A_{\mathrm{brand}}^{\mathrm{veh}}+A_{\mathrm{pair}}^{\mathrm{veh}}}{2}.
\end{aligned}
\label{eq:brand_vehicle}
\end{equation}

Here, $V$ is the set of complete vehicle groups, and overbars denote the voted brand and pair labels. The recognition parser extracts the first brace-delimited JSON object and reads its \texttt{brand} and \texttt{model} fields. Values are converted to strings before normalization; missing fields become empty strings, and extra fields are allowed. A returned view that cannot be parsed contributes empty labels to voting, whereas missing expected views make the vehicle group incomplete.

\subsection{Numerical Estimation and MRA}
\label{app:mra}

Numerical components configured for mean relative accuracy use absolute relative error for positive reference values. Accuracy is averaged over thresholds from 5\% to 50\% in 5\% increments:
\begin{equation}
s_{\mathrm{MRA}}(\hat y,y)=\frac{100}{10}\sum_{k=1}^{10}
\mathbb{1}\!\left[\frac{|\hat y-y|}{y}\leq 0.05k\right].
\label{eq:mra_detailed}
\end{equation}

MRA takes values in $\{0,10,\ldots,100\}$. After whitespace trimming, an SC response must contain only decimal digits and parse as a nonnegative integer; zero is a valid prediction. Count thresholds use integer arithmetic: threshold $k$ is satisfied when 100 times the absolute prediction error is no greater than $5kn$. Invalid responses score zero, and this comparison avoids floating-point tolerance. Vehicle dimensions must be finite positive values in millimeters; the scorer neither infers nor converts units from magnitude. LW averages length and width MRA, while HT scores height. In the frozen inventory, dimension records use a stored height proxy assigning them to 0--40 meters; they are excluded from the common-coverage height diagnostic.
\begin{equation}
100|\hat n-n|\leq 5kn.
\label{eq:count_integer}
\end{equation}

\subsection{Visual Grounding and Coordinate Conventions}
\label{app:grounding_scores}

A visual grounding response specifies one horizontal bounding box using its top-left and bottom-right coordinates. The default range is $[0,1000]$, with starting coordinates strictly smaller than ending coordinates along both axes. The primary metric is Acc@0.5:
\begin{equation}
\begin{aligned}
\operatorname{IoU}(\hat B,B)&=\frac{|\hat B\cap B|}{|\hat B\cup B|},\\
s_{\mathrm{VG}}&=100\,\mathbb{1}[\operatorname{IoU}(\hat B,B)\geq0.5].
\end{aligned}
\label{eq:grounding_detailed}
\end{equation}

The component score averages primary scores over its questions. This rule applies to SR, the nearest and farthest RD components, CG, and RI. Cross-view grounding is evaluated in the specified target image. Raw IoU may be retained diagnostically but does not contribute to the primary score.

Declared model coordinate conventions receive fixed conversions: GeoChat coordinates in $[0,100]$ are multiplied by ten; GeoLLaVA pixel coordinates are mapped to $[0,1000]$ using target-image width $W$ and height $H$. Box validity is checked in the original convention before conversion. Invalid boxes are not clipped, reordered, or repaired by extracting coordinates from explanatory text:
\begin{equation}
x'=\frac{1000x}{W},\qquad y'=\frac{1000y}{H}.
\label{eq:coordinate_conversion}
\end{equation}

\subsection{Angle and Geographic-Location Scoring}
\label{app:angle_gps}

Angular error uses the shortest periodic difference only when the question explicitly configures a period $P$; otherwise, it uses absolute difference. Let $\Delta$ be the predicted angle minus the reference angle:
\begin{equation}
e_{\theta}=
\begin{cases}
\min(\Delta\bmod P,\,P-(\Delta\bmod P)),
   &\text{$P$ set},\\
|\Delta|,&\text{otherwise}.
\end{cases}
\label{eq:angular_error}
\end{equation}
\begin{equation}
s_{\mathrm{angle}}=\frac{100}{3}
\sum_{\tau\in\{5^\circ,10^\circ,15^\circ\}}
\mathbb{1}[e_{\theta}\leq\tau+10^{-12}].
\label{eq:angle_score}
\end{equation}

Modulo normalization is applied only when enabled in the question configuration. When enabled, finite predictions are reduced modulo the configured period, and the original numeric bounds no longer apply to that periodic component. Each task's configuration specifies its period, legal range, and normalization flag.

GPS responses contain finite latitude and longitude values within $[-90,90]$ and $[-180,180]$, respectively. After conversion to radians, prediction and reference are compared using Haversine distance:
\begin{equation}
\begin{aligned}
a={}&\sin^2\!\left(\frac{\hat\phi-\phi}{2}\right)\\
&+\cos\phi\cos\hat\phi\,
\sin^2\!\left(\frac{\hat\lambda-\lambda}{2}\right).
\end{aligned}
\label{eq:haversine_a}
\end{equation}
\begin{equation}
d_{\mathrm{geo}}=2R\arcsin\!\left(\min\{1,\sqrt{\max\{0,a\}}\}\right).
\label{eq:haversine_d}
\end{equation}

The Earth-radius constant is $R=6{,}371{,}008.8$ meters. GPS scoring averages accuracies at distance thresholds of 1, 10, and 100 kilometers to obtain a 0--100 primary score:
\begin{equation}
s_{\mathrm{GPS}}=\frac{100}{3}
\sum_{\tau\in\{1000,10000,100000\}}
\mathbb{1}[d_{\mathrm{geo}}\leq\tau].
\label{eq:gps_score}
\end{equation}

\subsection{Exact Matching and Output Parsing}
\label{app:parsing}

For discrete text, the scorer trims surrounding whitespace, checks membership in the question's candidate set, and awards 100 only for an exact reference match. Comparison is case-sensitive and does not substitute synonyms. Numerical components configured for exact scoring receive 100 when the configured absolute or shortest-period error is at most $10^{-12}$, and zero otherwise.

Except for recognition-specific JSON extraction, numerical responses must match the required integer or numeric format in their entirety and satisfy configured legal ranges. Decimal and scientific notation are supported; nonfinite values are rejected. Horizontal-box and GPS responses must parse entirely as JSON arrays of four and two finite values, respectively. The scorer does not extract numbers from explanations, remove units, use an LLM judge, or repair partial formats.

\subsection{Aggregation and Prediction Coverage}
\label{app:aggregation}

Let $D_{t,c}$ be the observation set for component $c$ of task $t$, and let $s_i$ be an observation's primary score. Observations are pooled across evaluated scenes within a component, then averaged over components, tasks, and capability groups:
\begin{equation}
\begin{aligned}
S_{t,c}&=\frac{1}{|D_{t,c}|}\sum_{i\in D_{t,c}}s_i,\\
T_t&=\frac{1}{|C_t|}\sum_{c\in C_t}S_{t,c},\\
G_g&=\frac{1}{|\mathcal T_g|}\sum_{t\in\mathcal T_g}T_t,\\
S_{\mathrm{overall}}&=\frac{1}{4}\sum_{g=1}^{4}G_g.
\end{aligned}
\label{eq:aggregation_full}
\end{equation}

The four groups contain six, four, three, and four tasks, respectively, and each contributes 25\% to Overall. Each task therefore has weight $1/24$, $1/16$, $1/12$, or $1/16$, depending on its group. The fourth group is named viewpoint and context; it retains CP, PC, CL, and CH and the original weighting. CL and CH together contribute 12.5\% of Overall. Observation counts affect within-component averages but not the predefined equal weighting between components, tasks, and groups. BI and BV change both aggregation and statistical unit, so their difference is not an isolated estimate of a voting effect.

Returned but unparseable answers remain in the denominator and receive zero; recognition follows the empty-label rules above. Missing expected predictions prevent complete official scoring. In partial-coverage diagnostic mode, incomplete or empty components are unavailable, with this status propagated to tasks, groups, and Overall. Means over available responses alone are diagnostic. Duplicate or unknown question IDs are excluded from normal aggregation.

\subsection{Auxiliary Counting Statistics}
\label{app:counting_aux}

Let $D$ contain all count questions and $V$ the subset with valid numerical predictions. Auxiliary exact-match accuracy, valid-prediction mean absolute error (MAE), and root mean squared error (RMSE) are
\begin{equation}
\begin{aligned}
A_{\mathrm{count,exact}}
 &=\frac{1}{|D|}\sum_{i\in D}\mathbb{1}[i\in V\land\hat n_i=n_i],\\
\mathrm{MAE}_{\mathrm{valid}}
 &=\frac{1}{|V|}\sum_{i\in V}|\hat n_i-n_i|,\\
\mathrm{RMSE}_{\mathrm{valid}}
 &=\sqrt{\frac{1}{|V|}\sum_{i\in V}(\hat n_i-n_i)^2}.
\end{aligned}
\label{eq:counting_aux}
\end{equation}

Auxiliary exact-match accuracy uses a 0--1 scale. Invalid answers count as incorrect for this accuracy and are excluded from MAE and RMSE, with their number reported separately. MAE and RMSE are unavailable when no valid predictions exist. These auxiliary statistics do not contribute to the primary aggregate.

\section{Production Costs and Comparison Records}
\label{app:cost}

\subsection{Scene Capture Intervals}
\label{app:capture_time}
\begin{figure}[t]
\centering
\includegraphics[width=\columnwidth]{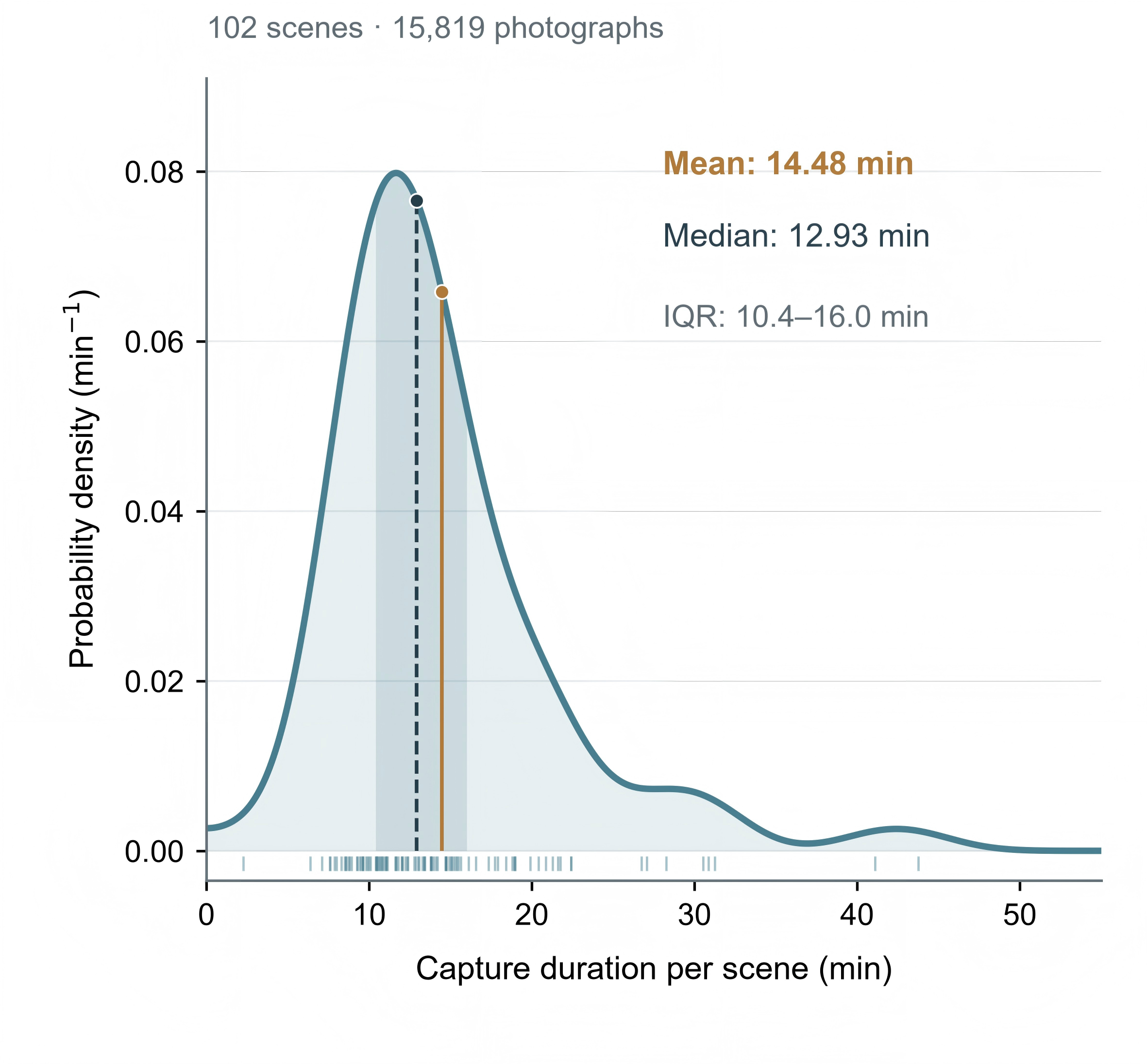}
\caption{Distribution of recorded capture intervals for 102 scenes. The Gaussian kernel density estimate uses Scott bandwidth (2.68 min) and reflection at zero.}
\label{fig:capture_intervals}\label{fig:capture_time_kde}
\end{figure}

For each of the 102 scenes in the acquisition inventory, the capture interval runs from the first to the last retained image, excluding travel, setup, and teardown. Across 15,819 images, the mean interval is 14.48 minutes, the median is 12.93 minutes, and the interquartile range is 10.4--16.0 minutes. This measures elapsed photography time, not labor or the complete production cycle. The observed distribution is shown in Figure~\ref{fig:capture_intervals}.

\subsection{Historical Annotation and Instance-Linked Correction Logs}
\label{app:annotation_timing}
The 11 historical scenes used image-by-image annotation in 2023, with a recorded mean of 136.31 minutes per scene. Instance-linked annotation began in 2026. July, August, and September contribute 7, 26, and 52 scene--month records, with means of 44.33, 34.80, and 29.52 minutes, respectively. A scene may contribute records in multiple months; the 85 current-period records are scene--month records. Together with the 11 historical records, the period-wise total is 96.

The current correction-log convention sums eligible operation intervals for a scene within each calendar month. When adjacent save events assigned to the same scene are more than five minutes apart, the interval is excluded; eligible intervals also exclude page reloads, different-scene access, and unassigned saves. Work outside these intervals is not measured.

The September recorded mean is 78.3\% below the historical mean; the displayed time-per-image summaries are 13.51 and 144.40 seconds, respectively. Because the periods use different measurement conventions and may differ in scene composition, observation count, and annotator experience, this descriptive comparison is not a causal estimate of workflow speedup.

\section{Rolling Evaluation and Historical Comparisons}
\label{app:stability}

\subsection{Disjoint Scene-Group Agreement}
\label{app:scene_group_consistency}
Using Start-10 and all 30 models, we form unordered pairs of disjoint $k$-scene subsets for $k=1,\ldots,5$. The number of pairs is
\[
N_k=\frac{1}{2}\binom{10}{k}\binom{10-k}{k},
\]
which gives 45, 630, 2,100, 1,575, and 126 comparisons. For each pair, Pearson and Spearman correlations use the shared model--task scores, excluding tasks absent from either subset. Table~\ref{tab:window_sizes} reports their arithmetic mean and sample standard deviation. Many subset pairs reuse scenes, so estimates are dependent; the standard deviations describe variation across sampled subsets rather than inferential uncertainty.

Pooling model--task entries can reflect differences in task difficulty as well as model capability. Consequently, these correlations describe agreement of score patterns, not the rank correlation of the 30-model Overall leaderboard. They are also distinct from the chronological, overlapping-window analysis in Fig.~\ref{fig:adjacent_consistency}.

\subsection{Capture-Time-Ordered Window Replay}
\label{app:window_construction}
The ten Start-10 scenes are ordered by capture time, with a fixed scene-ID order for ties. For window size $k$, window $j$ contains scenes $j$ through $j+k-1$, giving $11-k$ windows and $10-k$ adjacent replacements. Each replacement removes the earliest scene and admits the next. At $k=4$, windows cover scenes 1--4 through 7--10 and adjacent windows share three scenes.

Figure~\ref{fig:adjacent_consistency} presents six transitions between these seven windows and single-scene comparisons from the corresponding positions in the capture sequence. Single-scene correlations use the 14--17 task types shared by the compared scenes, rather than full-coverage Overall. Models, task definitions, and scoring rules are held fixed. This retrospective replay simulates proposed monthly replacements; it is not a prospective leaderboard run.

\subsection{Ranks on Tasks Shared by 2023 and 2026}
\label{app:year_rankings}
History-11 and Start-10 share six 2D tasks, two 3D tasks, and three Viewpoint and Context tasks:
\[
\begin{split}
\mathcal T_{\mathrm{2D}}^{\cap}&=\{\mathrm{BI,BV,SC,SR,WU,VH}\},\\
\mathcal T_{\mathrm{3D}}^{\cap}&=\{\mathrm{LW,HT}\},\\
\mathcal T_{\mathrm{Context}}^{\cap}&=\{\mathrm{CP,CL,CH}\}.
\end{split}
\]
RD, AD, CC, CG, RI, and PC are excluded. Observations are pooled within task components; tasks are averaged within each represented group, and the three groups are weighted equally:
\[
S_{\cap}=\frac{G_{\mathrm{2D}}^{\cap}+G_{\mathrm{3D}}^{\cap}
                         +G_{\mathrm{Context}}^{\cap}}{3}.
\]
Individual common 2D, 3D, and context tasks consequently have weights $1/18$, $1/4$, and $1/9$. No zero-valued cross-view group is inserted. This common-task aggregate is distinct from both the 17-task Overall and an unweighted mean of the 11 tasks.

\begin{figure}[t]
\centering
\includegraphics[width=\columnwidth]{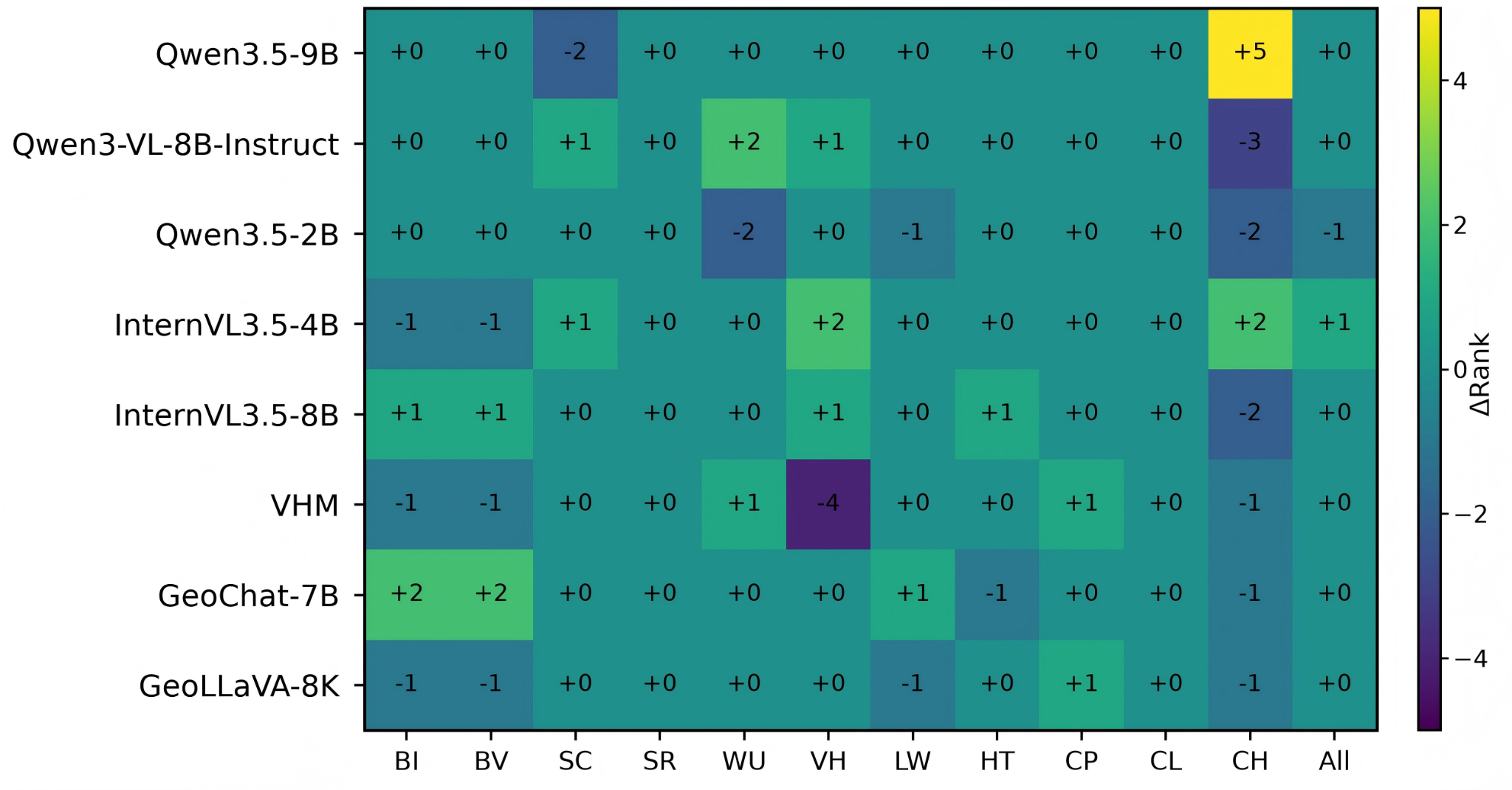}
\caption{Cross-year rank changes on 11 common tasks. Positive $\Delta\mathrm{Rank}=\mathrm{Rank}_{2023}-\mathrm{Rank}_{2026}$ denotes improvement. All is the equal-three-group common-task aggregate, not the full 17-task Overall.}
\label{fig:cross_year_ranks_appendix}
\end{figure}

The eight reference systems are Qwen3.5-9B, Qwen3-VL-8B-Instruct, Qwen3.5-2B, InternVL3.5-4B, InternVL3.5-8B, VHM, GeoChat-7B, and GeoLLaVA-8K. The historical comparison uses consistent model configurations across the two acquisition periods. Figure~\ref{fig:cross_year_ranks_appendix} uses $\Delta\mathrm{Rank}=\mathrm{Rank}_{2023}-\mathrm{Rank}_{2026}$, so positive values indicate a higher rank on Start-10. Tied scores share a rank.

Six systems retain their aggregate rank; Qwen3.5-2B moves down one position and InternVL3.5-4B moves up one. Task-level changes are larger: capture-hour rank changes range from $-3$ to $+5$, while VHM's heading rank changes by $-4$. Rank movement can also result from changes in ties. These results describe shared-task ranks; absolute difficulty and annotation quality were not calibrated across periods.

\begin{table*}[t]
  \centering
  \caption{Full task-level results for the 30 models on Start-10 full, reported on a 0--100 scale. Bold indicates column maxima, including ties. The four capability-group means are equally weighted for Overall.}
  \label{tab:full_task_results}\label{tab:main_results_unified_compatible}
  \begingroup
  \scriptsize
  \setlength{\tabcolsep}{2pt}
  \renewcommand{\arraystretch}{1.2}
  \resizebox{\textwidth}{!}{%
    \begin{tabular}{@{}l r *{7}{r}@{\hspace{6pt}}*{5}{r}@{\hspace{6pt}}*{4}{r}@{\hspace{6pt}}*{5}{r}@{}}
      \toprule[0.9pt]
      \multirow{2}{*}{Model} & \multirow{2}{*}{Overall} & \multicolumn{7}{c}{2D Perception} & \multicolumn{5}{c}{3D Perception} & \multicolumn{4}{c}{Cross-View Perception} & \multicolumn{5}{c}{Reasoning} \\
      \cmidrule(lr){3-9}\cmidrule(lr){10-14}\cmidrule(lr){15-18}\cmidrule(l){19-23}
       & & BI & BV & SC & SR & WU & VH & Mean & LW & HT & RD & AD & Mean & CC & CG & RI & Mean & CP & PC & CL & CH & Mean \\
      \midrule[0.55pt]
      \multicolumn{23}{l}{\textbf{Remote Sensing MLLMs}} \\
      GeoLLaVA-8K & 0.11 & 1.25 & 1.23 & 0.08 & 0.00 & 0.00 & 0.00 & 0.43 & 0.00 & 0.00 & 0.00 & 0.00 & 0.00 & 0.00 & 0.00 & 0.00 & 0.00 & 0.00 & 0.00 & 0.00 & 0.00 & 0.00 \\
      VHM & 4.38 & 1.08 & 0.77 & 30.62 & 0.00 & 20.00 & 6.54 & 9.83 & 0.06 & 1.25 & 0.00 & 0.00 & 0.33 & 19.43 & 0.00 & 0.00 & 6.48 & 0.80 & 2.77 & 0.00 & 0.00 & 0.89 \\
      GeoChat-7B & 8.31 & 1.86 & 2.00 & 25.40 & 0.58 & 20.00 & 1.91 & 8.63 & 1.07 & 13.00 & 0.00 & 14.53 & 7.15 & 22.91 & 0.00 & 0.00 & 7.64 & 17.66 & 11.41 & 0.00 & 10.21 & 9.82 \\
      \midrule
      \multicolumn{23}{l}{\textbf{Embodied MLLMs}} \\
      SenseNova-SI-1.1-InternVL3-2B & 9.13 & 5.00 & 4.77 & 36.73 & 18.19 & 46.67 & 2.55 & 18.99 & 0.10 & 0.05 & 7.02 & 0.05 & 1.81 & 24.14 & 3.61 & 1.34 & 9.69 & 4.24 & 4.84 & 3.40 & 11.67 & 6.04 \\
      Pelican1.0-VL-7B (2K) & 11.92 & 6.73 & 6.77 & 50.40 & 0.00 & 54.58 & 5.74 & 20.70 & 0.01 & 0.00 & 0.00 & 6.38 & 1.60 & 33.01 & 0.00 & 0.00 & 11.00 & 18.40 & 13.73 & 22.85 & 2.50 & 14.37 \\
      RynnBrain-2B & 14.08 & 7.71 & 9.08 & 54.79 & 59.69 & 26.46 & 2.71 & 26.74 & 0.11 & 0.00 & 33.77 & 0.96 & 8.71 & 35.08 & 0.00 & 0.00 & 11.69 & 3.64 & 6.00 & 0.00 & 27.08 & 9.18 \\
      Cosmos-Reason2-2B & 14.26 & 7.73 & 5.23 & 36.91 & 52.65 & 37.50 & 8.29 & 24.72 & 0.00 & 0.00 & 37.28 & 2.43 & 9.93 & 32.23 & 3.54 & 4.80 & 13.52 & 13.58 & 7.78 & 0.21 & 13.96 & 8.88 \\
      RynnBrain1.1-2B & 14.52 & 6.09 & 4.31 & 38.52 & 21.31 & 33.12 & 6.70 & 18.34 & 4.16 & 8.83 & 14.47 & 3.05 & 7.63 & 40.19 & 12.76 & 12.00 & 21.65 & 15.64 & 3.52 & 0.00 & 22.71 & 10.46 \\
      SenseNova-SI-1.1-InternVL3-8B & 14.79 & 4.83 & 5.54 & 47.36 & 42.17 & 58.12 & 3.35 & 26.89 & 0.00 & 15.95 & 11.84 & 0.03 & 6.95 & 31.18 & 6.66 & 0.42 & 12.75 & 12.59 & 11.62 & 3.68 & 22.29 & 12.55 \\
      ACE-Brain-0.5-8B & 14.95 & 11.13 & 16.00 & 58.56 & 1.31 & 56.67 & 6.70 & 25.06 & 2.11 & 13.28 & 5.70 & 19.55 & 10.16 & 33.46 & 0.00 & 0.14 & 11.20 & 22.65 & 11.46 & 6.53 & 12.92 & 13.39 \\
      SenseNova-SI-1.3-InternVL3-8B & 16.55 & 4.21 & 4.31 & 49.96 & 27.17 & 33.96 & 3.35 & 20.49 & 5.12 & 41.99 & 14.04 & 0.92 & 15.52 & 34.01 & 5.20 & 2.19 & 13.80 & 14.47 & 9.59 & 11.74 & \textbf{29.79} & 16.40 \\
      SenseNova-SI-1.5-InternVL3-8B & 16.88 & 3.61 & 2.92 & 46.09 & 53.43 & 27.92 & 3.51 & 22.91 & 0.00 & 0.62 & 31.14 & 7.89 & 9.91 & 35.75 & 12.00 & 11.08 & 19.61 & 21.30 & 5.50 & 5.76 & 27.71 & 15.07 \\
      SenseNova-SI-1.4-InternVL3-8B & 18.95 & 4.48 & 4.77 & 57.04 & 43.41 & 20.21 & 3.03 & 22.16 & 0.00 & 0.00 & 41.23 & 6.80 & 12.01 & 38.05 & 17.06 & 14.96 & 23.36 & 26.47 & 14.43 & 15.00 & 17.29 & 18.30 \\
      Cosmos-Reason2-8B & 20.80 & 7.49 & 7.85 & 38.94 & 68.48 & 30.42 & 8.93 & 27.02 & 16.02 & 4.77 & 48.25 & 0.64 & 17.42 & 34.88 & 2.77 & 3.46 & 13.71 & 38.76 & 14.85 & 16.81 & \textbf{29.79} & 25.05 \\
      ACE-Brain-0-8B & 21.15 & 7.67 & 8.46 & 42.87 & 38.18 & 22.71 & 2.87 & 20.46 & 33.77 & 45.74 & 12.28 & 39.15 & 32.74 & 33.32 & 0.00 & 0.64 & 11.32 & 39.26 & 19.73 & 5.56 & 15.83 & 20.09 \\
      MiMo-Embodied-7B & 23.98 & 6.32 & 8.46 & 55.69 & 0.00 & 57.71 & 13.88 & 23.68 & 79.27 & 66.99 & 0.00 & 22.04 & 42.07 & 35.60 & 0.00 & 0.00 & 11.87 & 43.11 & 5.17 & 17.22 & 7.71 & 18.30 \\
      PhysBrain1.5-8B & 24.09 & 10.44 & 14.31 & 51.98 & 1.74 & 71.67 & 5.74 & 25.98 & 43.12 & 72.22 & 1.75 & 15.40 & 33.13 & 41.06 & 4.85 & 12.07 & 19.33 & 21.12 & 15.67 & 5.56 & 29.38 & 17.93 \\
      RynnBrain-8B & 31.69 & 6.90 & 8.62 & 62.89 & 70.73 & 34.79 & 5.90 & 31.64 & 51.69 & 86.86 & \textbf{57.46} & 7.27 & 50.82 & 36.39 & 7.35 & 12.56 & 18.77 & 52.30 & 13.73 & 14.79 & 21.25 & 25.52 \\
      RoboBrain2.5-8B-NV & 32.84 & 8.39 & 9.69 & 56.36 & 64.49 & 60.83 & 13.56 & 35.55 & 64.79 & 85.02 & 41.67 & 38.48 & 57.49 & 36.40 & 3.05 & 8.82 & 16.09 & 45.20 & 13.15 & 20.42 & 10.21 & 22.24 \\
      RynnBrain1.1-9B & 35.10 & 8.30 & 10.00 & 56.31 & 62.85 & 43.12 & 6.22 & 31.13 & 84.72 & 92.03 & 39.91 & 10.26 & 56.73 & 43.81 & 25.31 & 44.25 & 37.79 & 16.43 & 3.68 & 12.85 & 26.04 & 14.75 \\
      ACE-Brain-0.6-9B & 43.29 & 6.62 & 2.31 & 66.12 & 69.14 & 54.79 & 4.78 & 33.96 & 95.44 & \textbf{95.56} & 53.07 & 34.02 & 69.52 & 45.31 & 46.46 & 76.15 & 55.97 & 18.29 & 10.63 & 10.07 & 15.83 & 13.70 \\
      \midrule
      \multicolumn{23}{l}{\textbf{General-Purpose MLLMs}} \\
      Qwen3.5-2B & 15.16 & 12.03 & 14.62 & 42.25 & 66.30 & 60.83 & 6.70 & 33.79 & 0.48 & 4.22 & 42.11 & 6.48 & 13.32 & 10.11 & 4.30 & 3.74 & 6.05 & 14.32 & 6.00 & 4.37 & 5.21 & 7.48 \\
      InternVL3.5-8B & 17.55 & 4.79 & 4.77 & \textbf{66.91} & 0.00 & 55.62 & 12.92 & 24.17 & 53.67 & 20.85 & 0.00 & 8.41 & 20.73 & 43.40 & 0.00 & 0.00 & 14.47 & 26.93 & 4.14 & 1.39 & 10.83 & 10.82 \\
      InternVL3.5-4B & 18.81 & 4.73 & 4.92 & 56.63 & 0.00 & 43.33 & 10.21 & 19.97 & 70.01 & 40.89 & 0.00 & 0.03 & 27.73 & 34.69 & 0.00 & 0.00 & 11.56 & 30.12 & 20.97 & 0.97 & 11.88 & 15.98 \\
      Qwen3-VL-8B-Instruct & 31.16 & 11.83 & 17.23 & 63.47 & 73.67 & \textbf{76.88} & 10.05 & 42.19 & 50.12 & 88.72 & 48.68 & 4.46 & 47.99 & 33.37 & 9.29 & 3.95 & 15.54 & 26.80 & 15.34 & 22.85 & 10.62 & 18.90 \\
      Qwen3.5-9B & 36.08 & 13.87 & 20.31 & 57.47 & 74.72 & 68.12 & 6.70 & 40.20 & 96.67 & 91.83 & 51.75 & 36.98 & 69.31 & 27.14 & 2.01 & 17.93 & 15.69 & 28.46 & 10.38 & \textbf{25.69} & 11.88 & 19.10 \\
      DeepSeek-V4.1-Flash-1017 / low & 37.19 & 11.14 & 17.08 & 61.68 & 67.11 & 46.04 & 14.99 & 36.34 & 96.08 & 90.24 & 35.09 & 21.41 & 60.70 & 40.72 & 14.15 & 42.27 & 32.38 & 28.78 & 22.00 & 14.24 & 12.29 & 19.33 \\
      GPT-5.6-Luna-low & 37.77 & 8.57 & 8.00 & 50.59 & 46.20 & 54.38 & 33.65 & 33.56 & 95.62 & 91.63 & 14.91 & 42.83 & 61.25 & \textbf{51.14} & 12.69 & 32.04 & 31.96 & 50.15 & 19.93 & 17.50 & 9.58 & 24.29 \\
      GPT-6.1-Sol-low & 59.13 & 13.53 & 24.92 & 55.57 & \textbf{82.91} & 69.58 & \textbf{79.11} & \textbf{54.27} & 97.11 & 94.85 & 48.25 & 66.44 & 76.66 & 48.70 & 57.21 & 94.99 & 66.97 & 64.29 & 52.03 & 25.00 & 13.12 & 38.61 \\
      GPT-6-Astra-low & \textbf{59.88} & \textbf{13.98} & \textbf{26.31} & 57.18 & 82.51 & 67.71 & 77.35 & 54.17 & \textbf{97.38} & 95.36 & 46.49 & \textbf{70.60} & \textbf{77.46} & 49.28 & \textbf{61.10} & \textbf{96.82} & \textbf{69.07} & \textbf{66.74} & \textbf{53.31} & 25.21 & 10.00 & \textbf{38.81} \\
      \bottomrule[0.9pt]
    \end{tabular}%
  }
  \endgroup
\end{table*}

\section{Complete Model Evaluation Results}
\label{app:model_results}
The Start-10 full results cover 30 models, 35,491 question records, and 17 scoring tasks (Table~\ref{tab:full_task_results}). All systems evaluate the full collection. BI and BV share recognition records, with BV aggregated at vehicle level. These results describe Start-10; a future Rolling-4 leaderboard requires its own four-scene manifest.

\subsection{Evaluation Interface}
\label{app:model_interface}

\paragraph{Model-visible information.}
Only question text and images are supplied. Recognition uses vehicle crops; other tasks use original images without evaluator-added visual markers or external resizing/cropping. Multiple images are delivered separately and simultaneously. Embedded metadata are removed; original metadata remain available for reference construction and diagnostic grouping. Internal image tokenization or resizing can still differ between systems. In RI, the source instance is specified in the question using a point normalized to $[0,1000]$ in the source image, and the answer refers to the target image (Appendix~\ref{app:task_demos}).

\paragraph{Execution and answer validation.}
Retries apply to timeouts, network failures, and service errors. Returned empty answers, invalid JSON, and invalid formats receive zero. The reported runs cover every expected item, but returned predictions may still be invalid. Scores reflect both task performance and format compliance under the fixed parser; the aggregate does not separate these factors.

\begin{table*}[t]
  \centering
  \caption{Start-10 Full overall scores and single A100-40G GPU inference efficiency
  for 26 models. Generation speed was measured on 50 questions.}
  \label{tab:overall_inference_efficiency}
  \small
  \setlength{\tabcolsep}{8pt}
  \renewcommand{\arraystretch}{1.12}
  \begin{tabular}{@{}lrrr@{}}
    \toprule
    Model
      & Overall
      & Generation Speed (questions/s)
      & Peak GPU Memory (MiB) \\
    \midrule
    ACE-Brain-0.6-9B                    & 43.29 & 0.123 & 30,687 \\
    Qwen3.5-9B                          & 36.08 & 0.119 & 25,090 \\
    RynnBrain1.1-9B                     & 35.10 & 0.122 & 25,093 \\
    RoboBrain2.5-8B-NV                  & 32.84 & 1.478 & 17,335 \\
    RynnBrain-8B                        & 31.69 & 1.198 & 17,335 \\
    Qwen3-VL-8B-Instruct                & 31.16 & 0.167 & 24,474 \\
    PhysBrain1.5-8B                     & 24.09 & 0.161 & 24,737 \\
    MiMo-Embodied-7B                    & 23.98 & 0.263 & 20,652 \\
    ACE-Brain-0-8B                      & 21.15 & 1.355 & 16,787 \\
    Cosmos-Reason2-8B                   & 20.80 & 1.186 & 17,337 \\
    SenseNova-SI-1.4-InternVL3-8B       & 18.95 & 1.250 & 16,536 \\
    InternVL3.5-4B                      & 18.81 & 0.722 & 10,593 \\
    InternVL3.5-8B                      & 17.55 & 0.557 & 18,030 \\
    SenseNova-SI-1.5-InternVL3-8B       & 16.88 & 1.253 & 16,536 \\
    SenseNova-SI-1.3-InternVL3-8B       & 16.55 & 1.137 & 16,536 \\
    Qwen3.5-2B                          & 15.16 & 0.198 & 8,732  \\
    ACE-Brain-0.5-8B                    & 14.95 & 0.154 & 24,065 \\
    SenseNova-SI-1.1-InternVL3-8B       & 14.79 & 1.024 & 16,536 \\
    RynnBrain1.1-2B                     & 14.52 & 0.219 & 8,732  \\
    Cosmos-Reason2-2B                   & 14.26 & 1.870 & 4,473  \\
    RynnBrain-2B                        & 14.08 & 1.443 & 4,473  \\
    Pelican1.0-VL-7B (2K)              & 11.92 & 0.287 & 20,615 \\
    SenseNova-SI-1.1-InternVL3-2B       &  9.13 & 1.423 & 4,664  \\
    GeoChat-7B                          &  8.31 & 1.392 & 16,851 \\
    VHM                                 &  4.38 & 1.120 & 14,942 \\
    GeoLLaVA-8K                         &  0.11 & 0.903 & 24,412 \\
    \bottomrule
  \end{tabular}
\end{table*}

\subsection{Score Interpretation and Version Scope}
\label{app:model_interpretation}
GPT-6-Astra-low obtains 59.88 Overall, GPT-6.1-Sol-low obtains 59.13, and ACE-Brain-0.6-9B obtains 43.29. BI/BV combine brand and joint brand--model accuracy. Dimensions use specification-based references and MRA tolerances. Qwen3-VL-8B-Instruct, for example, scores 88.72 on HT and 4.46 on AD.

All 18 tested embodied models exceed every remote-sensing comparator in Overall, while group-level results vary: SenseNova-SI-1.1-InternVL3-2B scores 1.81 in 3D, below GeoChat-7B's 7.15, and Cosmos-Reason2-2B scores 8.88 in Viewpoint and Context, below GeoChat-7B's 9.82.

Predictions, reference answers, and inclusion conditions are versioned, and reference corrections require recomputation. The 12-model height/pitch heatmap in Appendix~\ref{app:geometry} reports a separate eight-task conditional macro-average. Model-family differences are descriptive comparisons among the evaluated systems.

\subsection{Inference Efficiency}
Table~\ref{tab:overall_inference_efficiency} reveals a clear trade-off between benchmark performance and inference efficiency. ACE-Brain-0.6-9B achieves the highest Overall score of 43.29, but requires the largest peak GPU memory (30,687\,MiB) and processes only 0.123 questions per second. Qwen3.5-9B and RynnBrain1.1-9B also obtain relatively high Overall scores, while exhibiting similarly low generation speeds and memory consumption of approximately 25\,GiB. In contrast, RoboBrain2.5-8B-NV and RynnBrain-8B provide more favorable accuracy--efficiency balances: they achieve Overall scores of 32.84 and 31.69 at generation speeds of 1.478 and 1.198 questions per second, respectively. Among the smaller models, Cosmos-Reason2-2B is the fastest model at 1.870 questions per second and uses only 4,473\,MiB of peak GPU memory, although its Overall score is limited to 14.26. These results indicate that parameter scale and computational cost do not translate monotonically into either higher benchmark performance or faster inference. Model selection should therefore consider the target deployment setting: larger, slower models may be preferred when accuracy is the primary objective, whereas RoboBrain2.5-8B-NV, RynnBrain-8B, and Cosmos-Reason2-2B offer stronger alternatives when throughput and memory efficiency are also important.
\section{Observation-Geometry Diagnostics}
\label{app:geometry}

\subsection{Diagnostic Definition and Coverage}
\label{app:geometry_coverage}
The diagnostic task set is $\mathcal T_{\mathrm{diag}}=\{\mathrm{BI,BV,WU,RD,CP,PC,CL,CH}\}$. Each task has equal weight, giving $D_b=\frac18\sum_{t\in\mathcal T_{\mathrm{diag}}}T_{t,b}$ for condition bin $b$. This differs from the four-group Overall. Conditions are assigned using the original capture metadata, which are excluded from model inputs. Returned invalid answers score zero.

BI and BV reuse 18,228 recognition views. BI evaluates individual observations; BV votes within physical vehicles. The height bins contain 324, 320, and 315 vehicle groups, with 325 unique vehicles overall. A vehicle can occur in several bins, so these counts must not be summed as independent entities. Table~\ref{tab:diagnostic_height_coverage} counts shared BI/BV input records only once in its total.

\begin{table}[t]
\centering
\caption{Diagnostic height coverage. Heights are in meters. BI and BV reuse the same recognition views; BV is scored over the vehicle groups shown separately.}
\label{tab:diagnostic_height_coverage}
\small
\setlength{\tabcolsep}{4pt}
\begin{tabular}{lrrr}
\toprule
Input or unit & $[0,40]$ & $(40,80]$ & $(80,120]$ \\
\midrule
BI/BV shared views & 4,360 & 5,789 & 8,079 \\
BV vehicle groups & 324 & 320 & 315 \\
WU records & 175 & 147 & 158 \\
RD records & 56 & 30 & 9 \\
CP records & 495 & 238 & 227 \\
PC records & 235 & 307 & 264 \\
CL records & 175 & 147 & 158 \\
CH records & 175 & 147 & 158 \\
\midrule
Distinct input records & 5,671 & 6,805 & 9,053 \\
\bottomrule
\end{tabular}
\end{table}

These counts yield 21,529 distinct input records; treating BI and BV as separate inputs would double-count 18,228 views and yield 39,757 task-input entries.

The diagnostic includes only the eight tasks listed above; SC has records in all height bins but is excluded. LW/HT records use stored height proxies assigned to the 0--40\,m bin rather than measured capture heights, so these tasks are omitted from the height analysis.

\subsection{Pitch Bins and Vehicle Voting}
\label{app:geometry_voting}
Pitch is the absolute downward angle, with bins at $30^\circ$, $45^\circ$, $60^\circ$, and $90^\circ$ and tolerance $\pm7.5^\circ$. Recognition-view counts are 4,543, 5,607, 5,547, and 2,490, respectively. Their total is 18,187; the remaining 41 recognition views fall outside the bins. Distinct vehicle counts are 325, 324, 322, and 283, giving 1,254 vehicle--bin incidences for 325 globally unique vehicles.

BV votes are local to each condition: (height bin, vehicle ID) or (pitch bin, vehicle ID). Predictions outside the condition are excluded. Brand and joint brand--model labels are voted separately, with lexicographic tie-breaking under Appendix~\ref{app:recognition_scores}. Vehicles can appear in multiple bins with different view counts, so bin estimates are dependent.





\section{Illustrative Examples for All 17 Tasks}
\label{app:task_demos}
The following cards show existing examples from the 928-question minival (labeled New10 MiniVal in the graphics). Each presents an image, a question, its human-verified reference, and selected model responses. Card-level counts concern the displayed systems and differ from the full 30-model evaluation.

The cards are presentation composites: reference answers, response panels, capture-metadata strips, and explanatory overlays are display annotations, not model inputs. Evaluation supplies only question text and image(s), with embedded metadata removed, as specified in Appendix~\ref{app:model_interface}.
\onecolumn
\newcommand{\taskdemocard}[2]{%
  \clearpage
  \noindent{\large\bfseries #1}\par\vspace{0.3em}
  \begin{center}
    \includegraphics[width=0.99\textwidth,height=0.84\textheight,keepaspectratio]{#2}
  \end{center}
}
\taskdemocard{BI: Brand/model recognition (per image)}{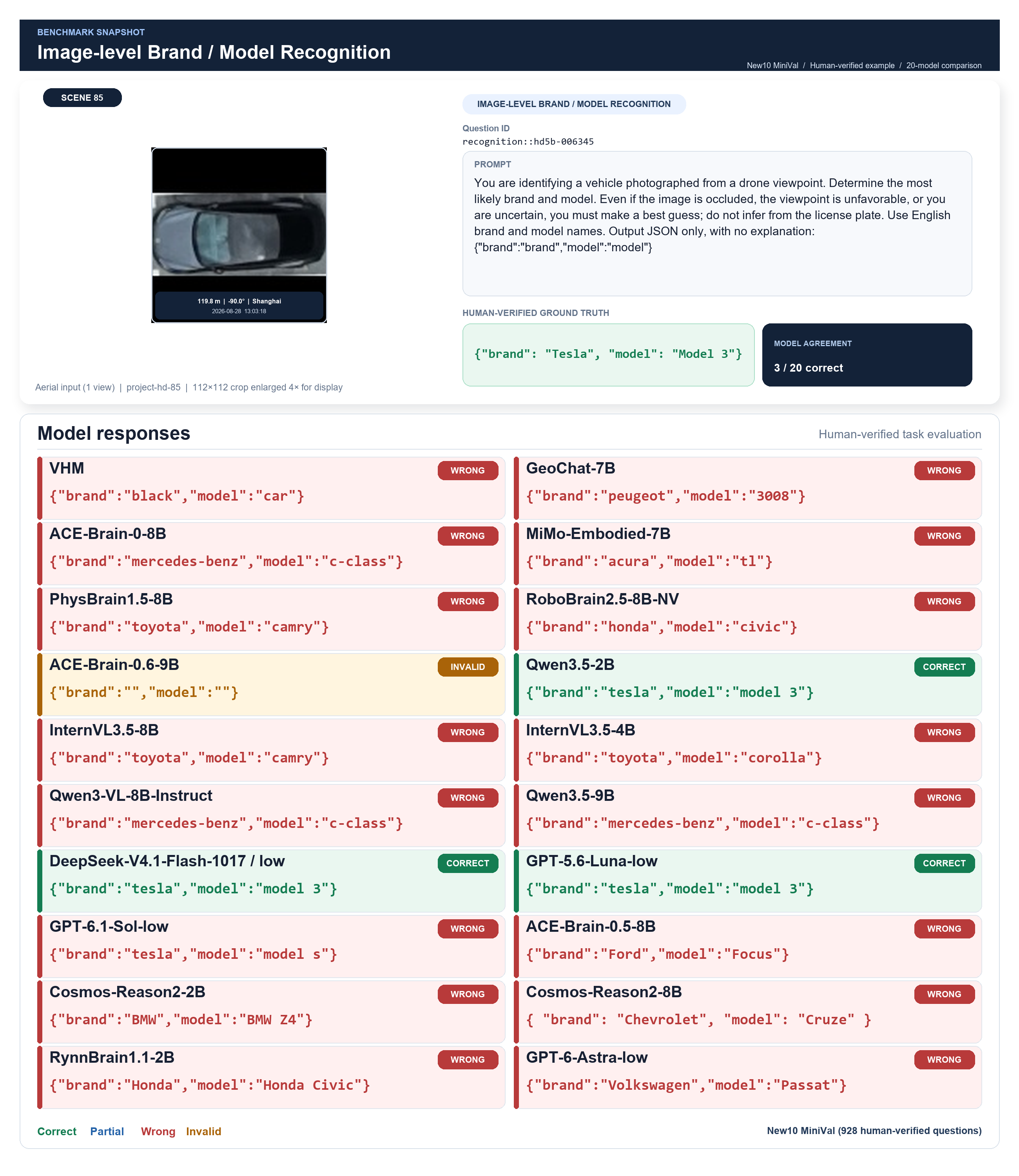}
\taskdemocard{BV: Brand/model recognition (per vehicle)}{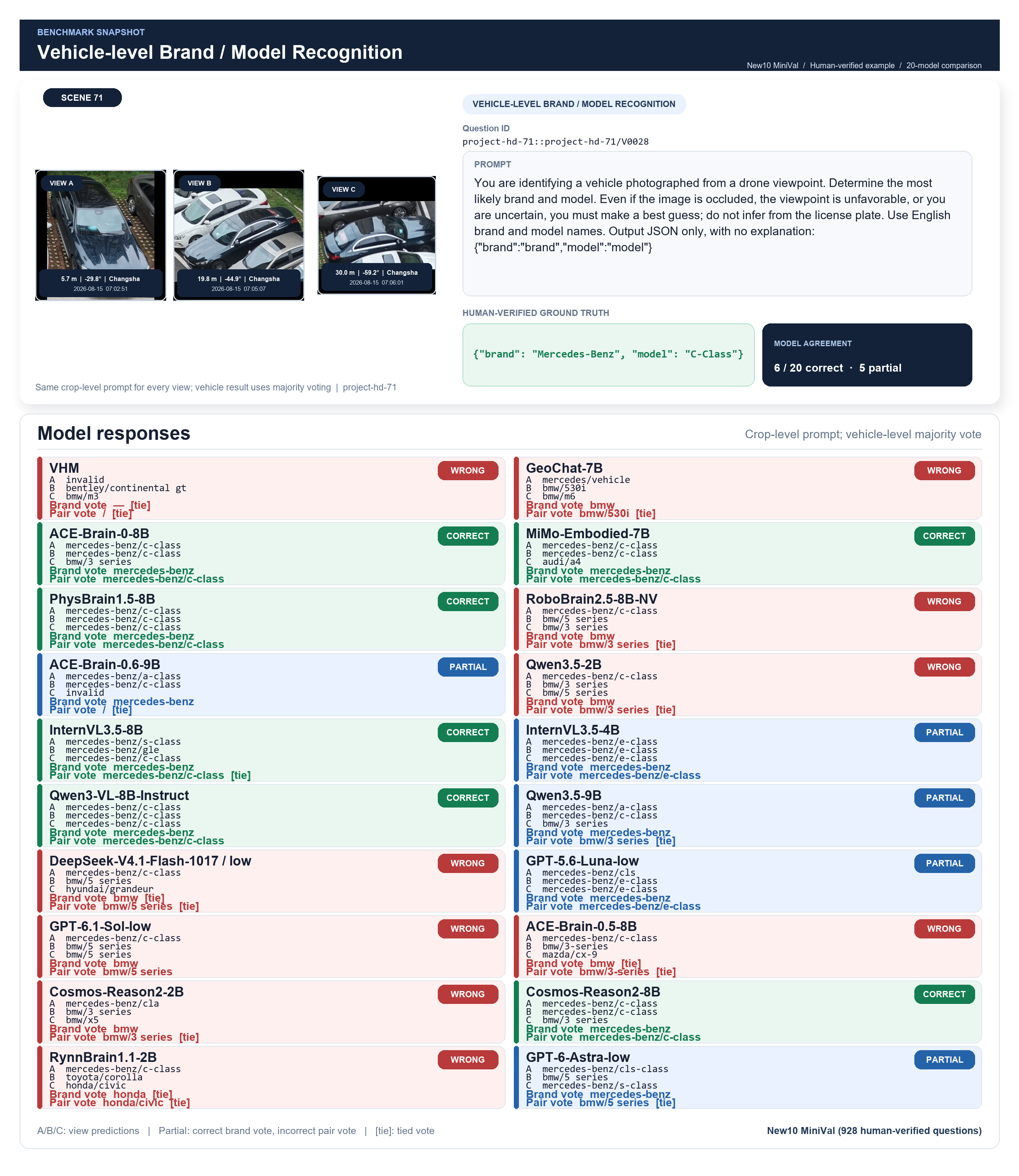}
\taskdemocard{SC: Single-image counting}{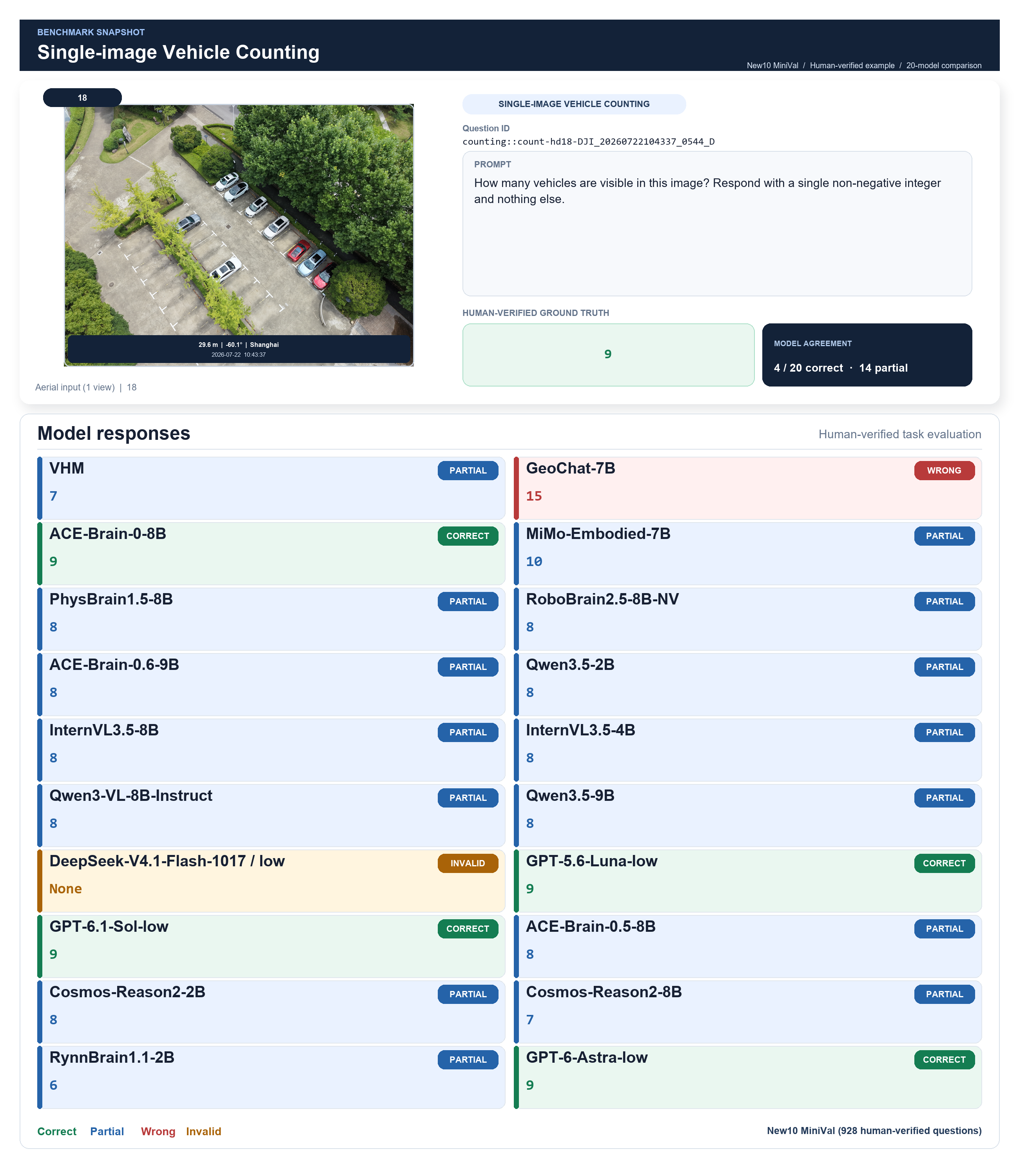}
\taskdemocard{SR: 2D spatial relations}{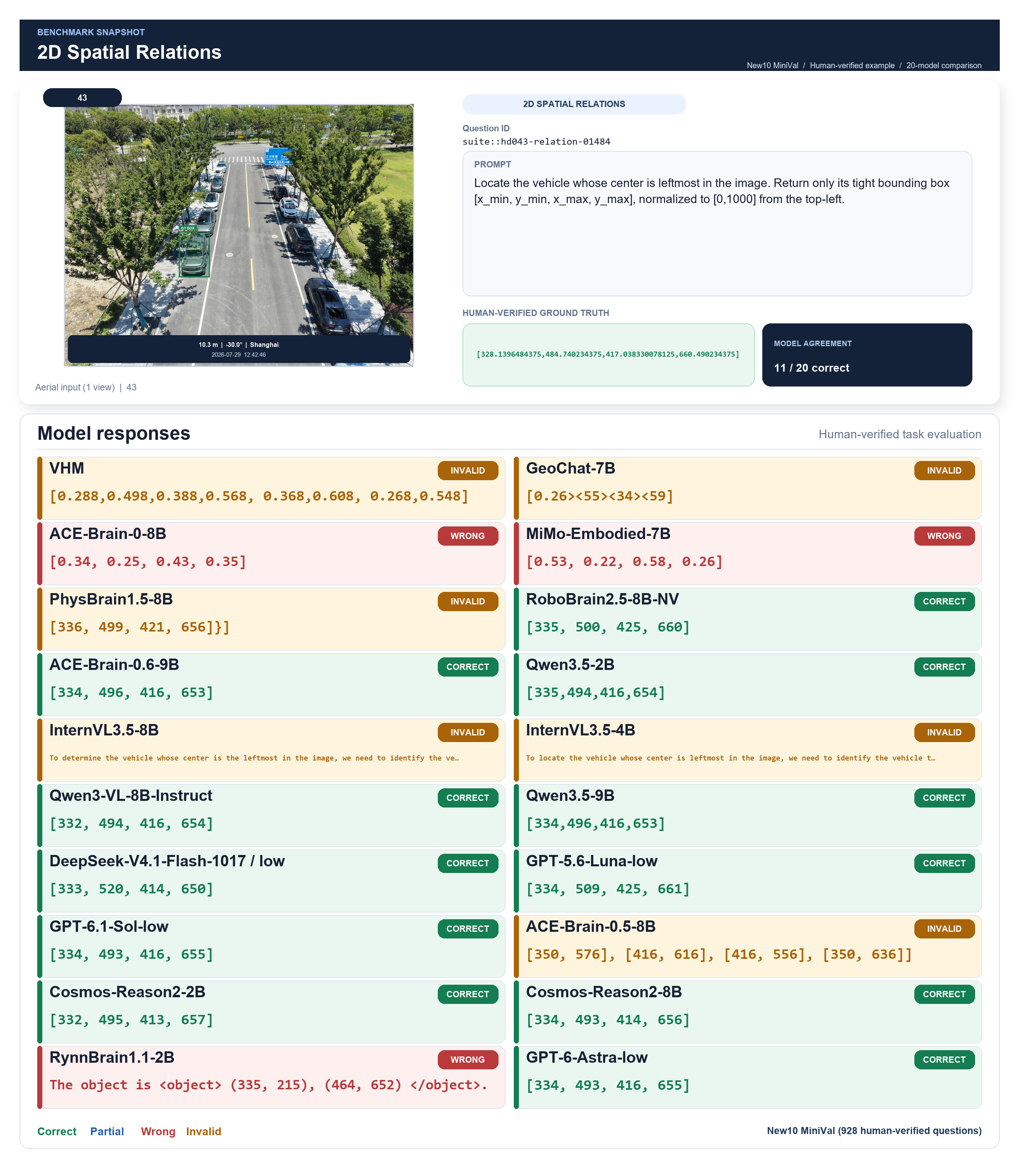}
\taskdemocard{WU: Weather understanding}{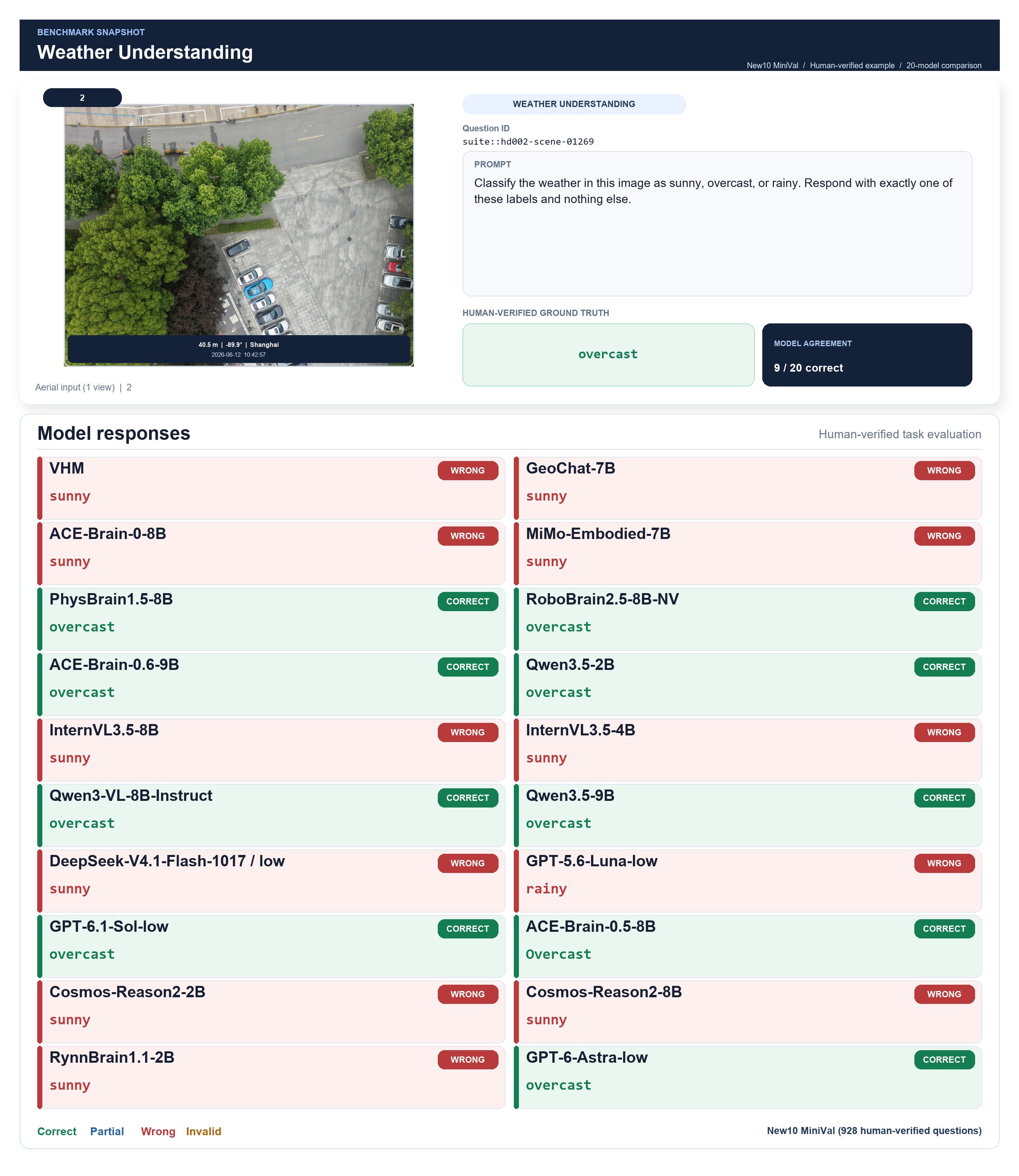}
\taskdemocard{VH: 2D vehicle heading}{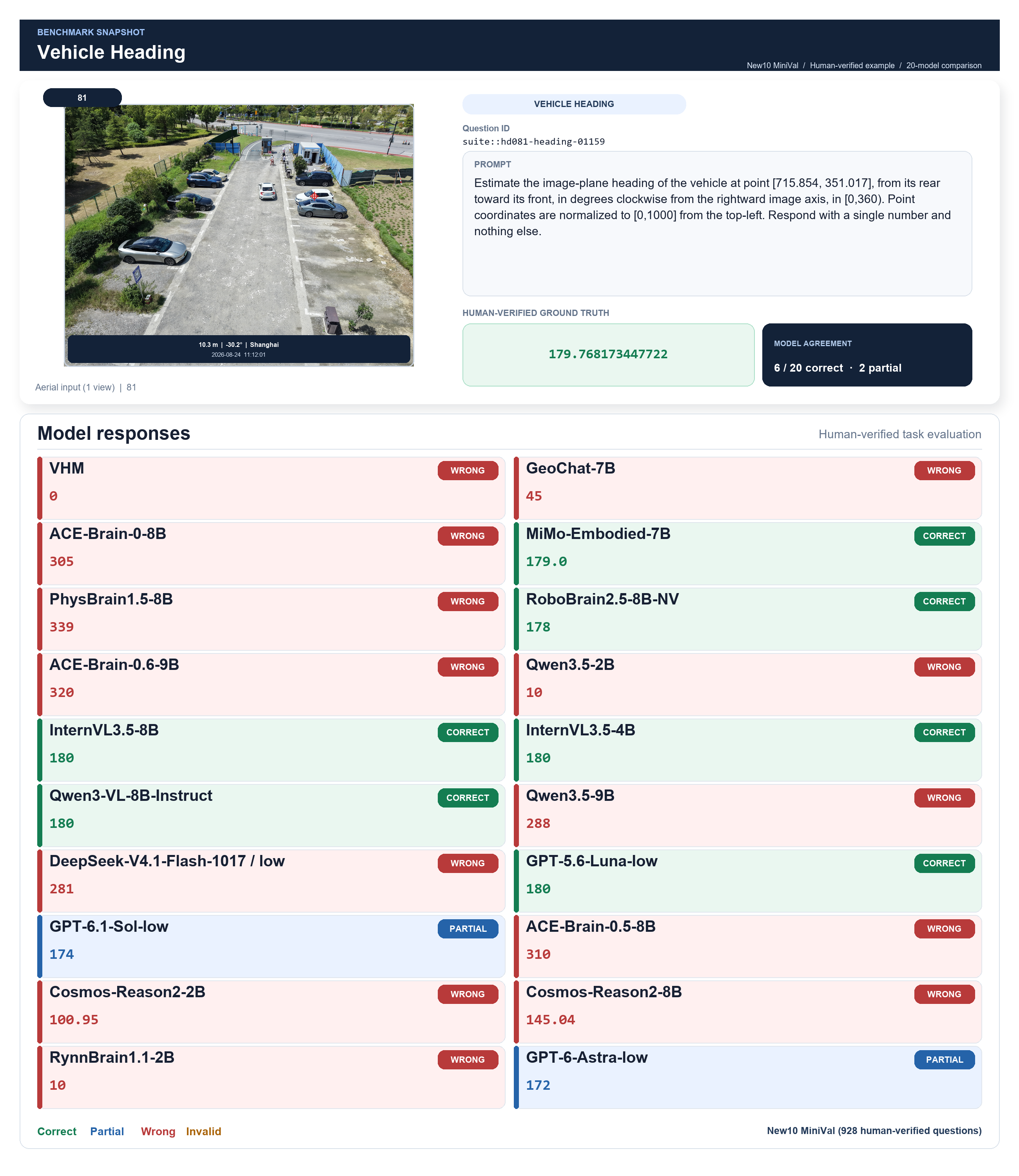}
\taskdemocard{LW: Vehicle length and width}{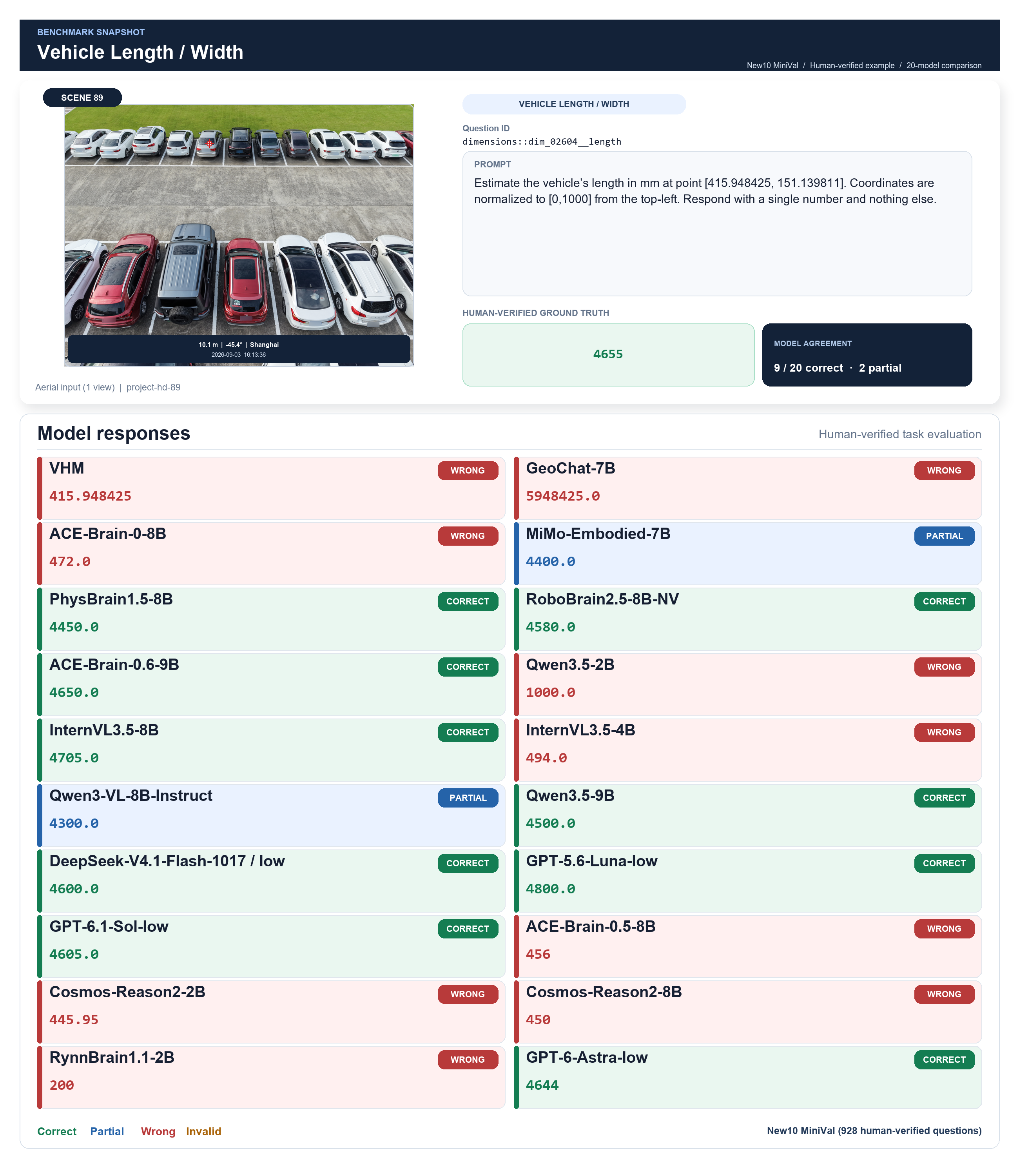}
\taskdemocard{HT: Vehicle height}{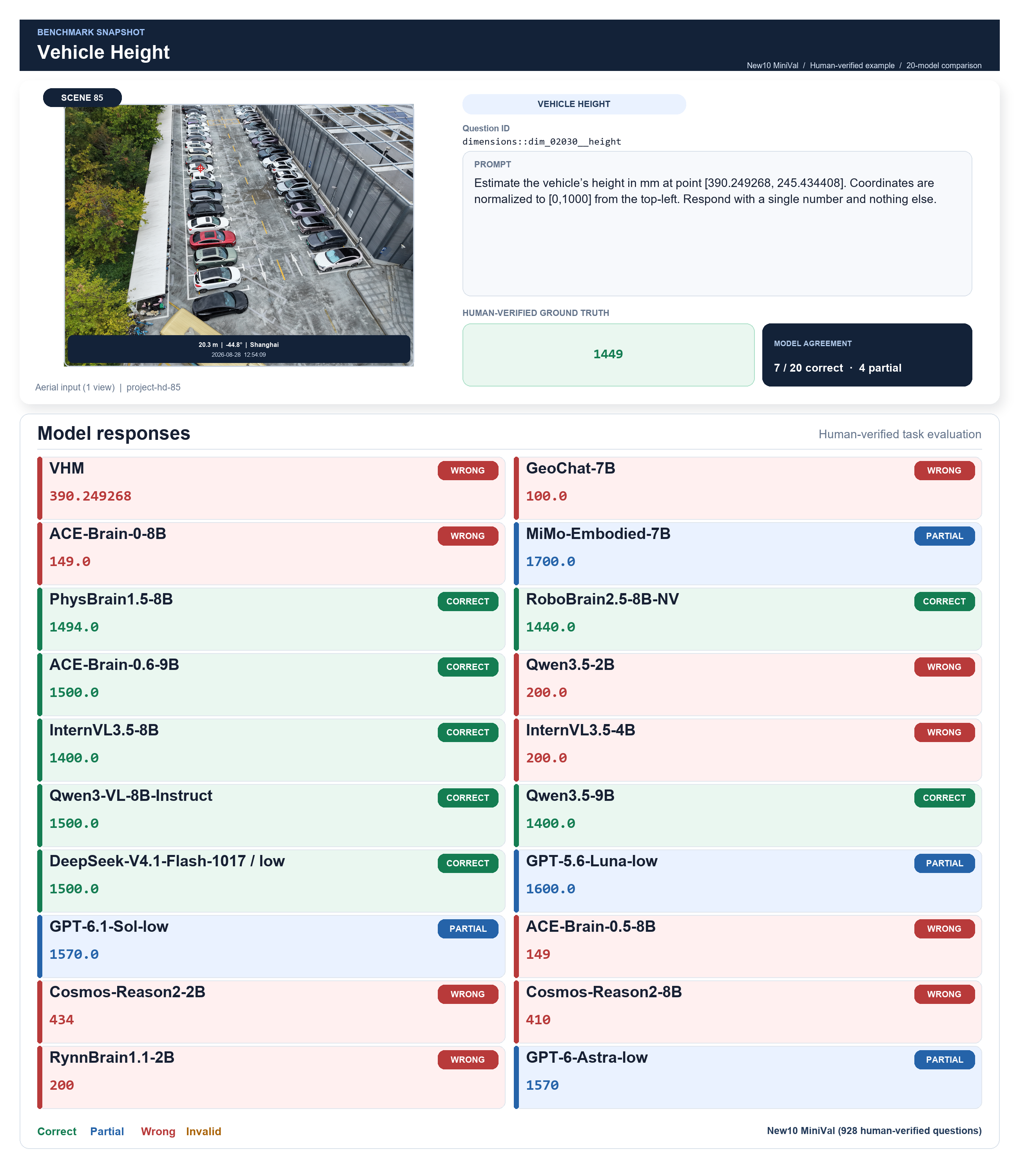}
\taskdemocard{RD: Relative spatial distance}{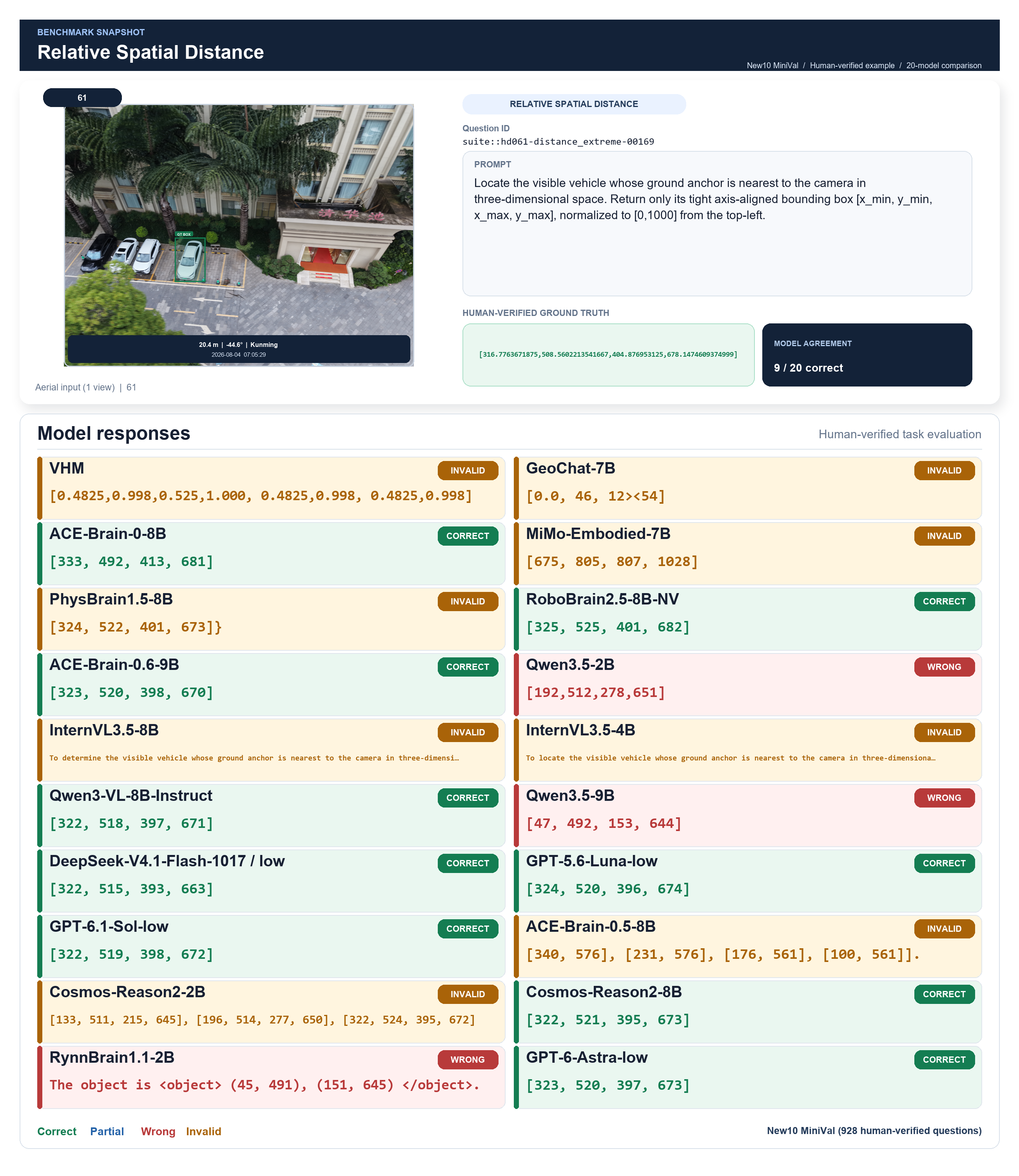}
\taskdemocard{AD: Absolute spatial distance}{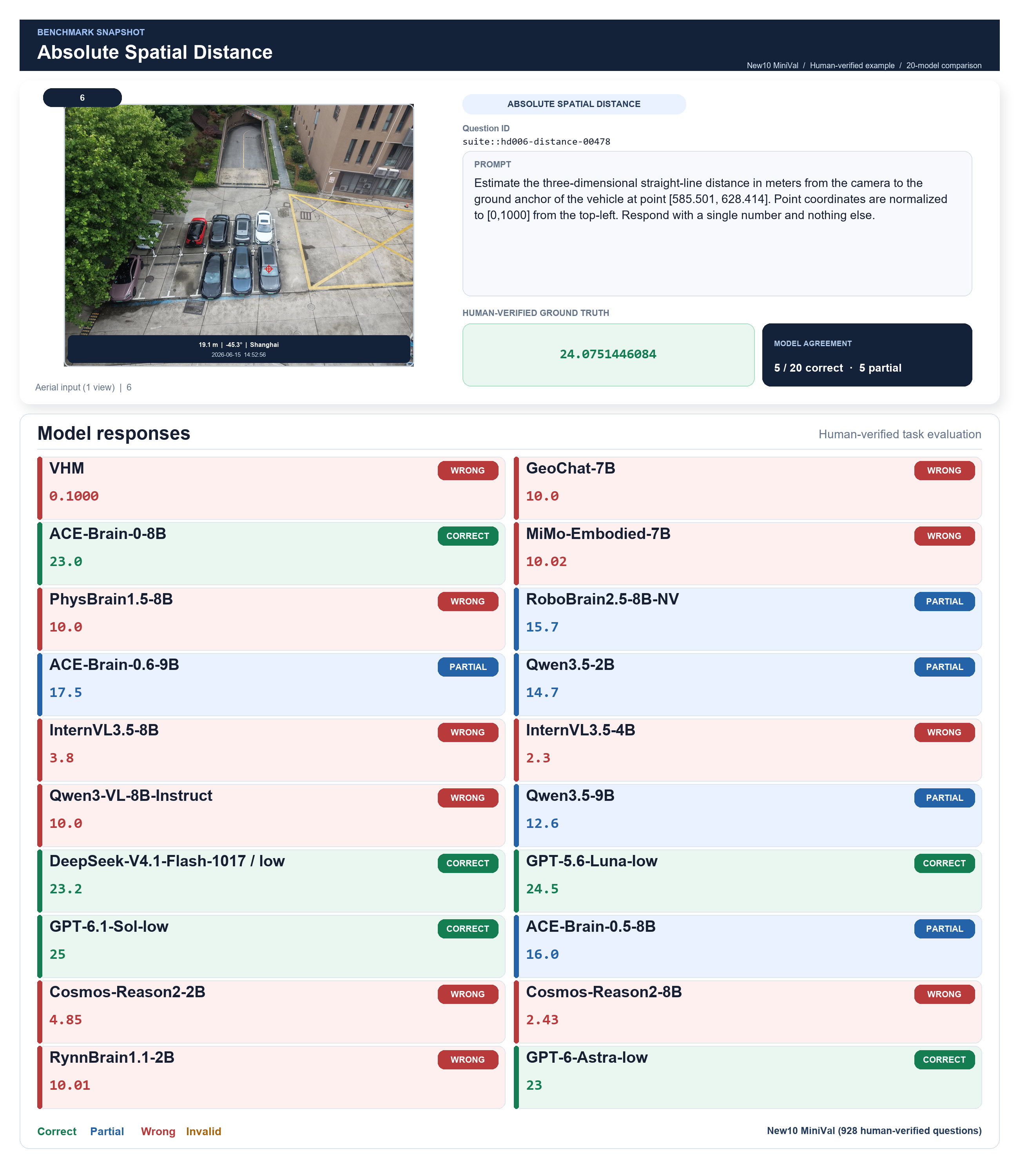}
\taskdemocard{CC: Cross-view counting}{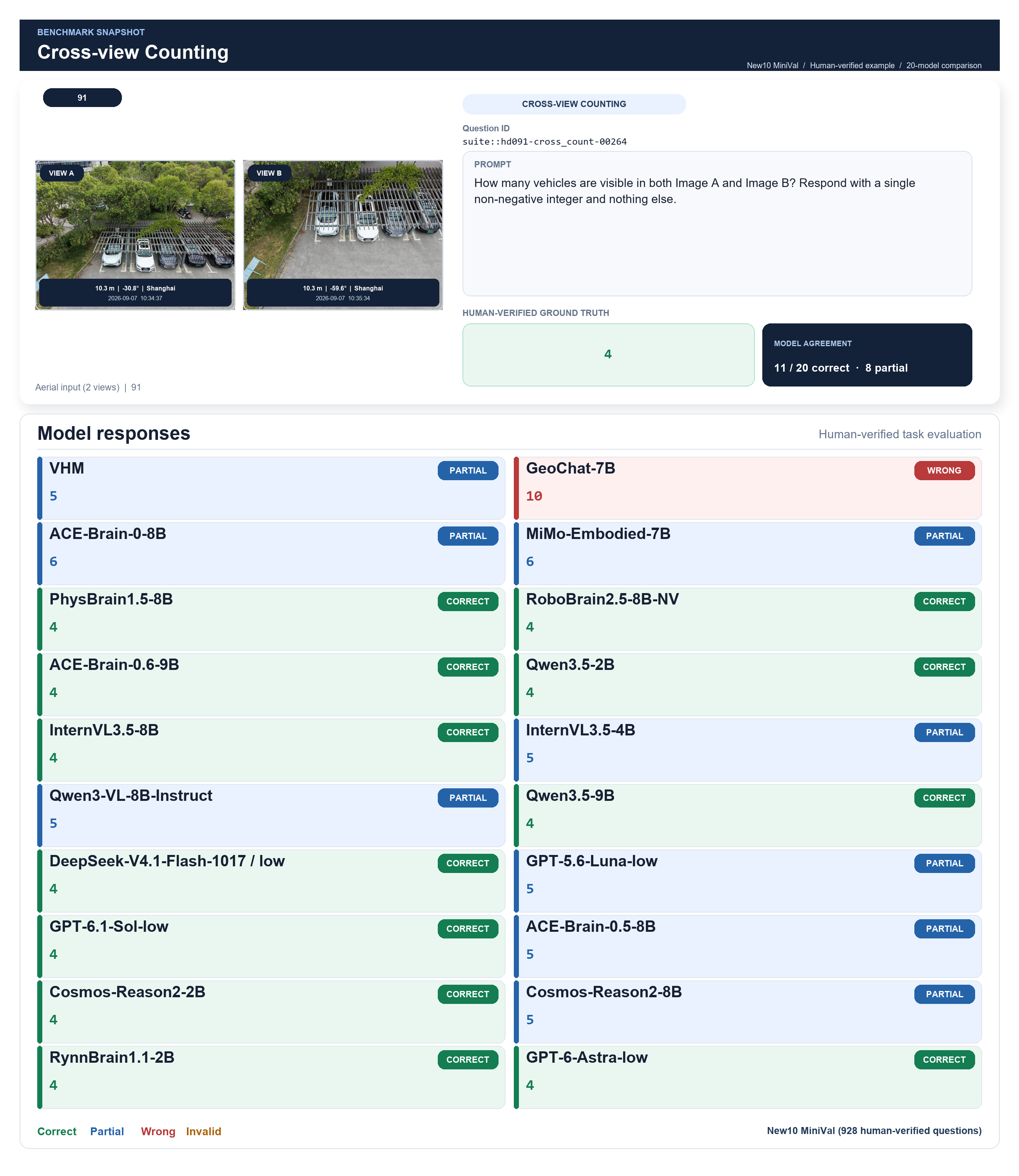}
\taskdemocard{CG: Cross-view grounding}{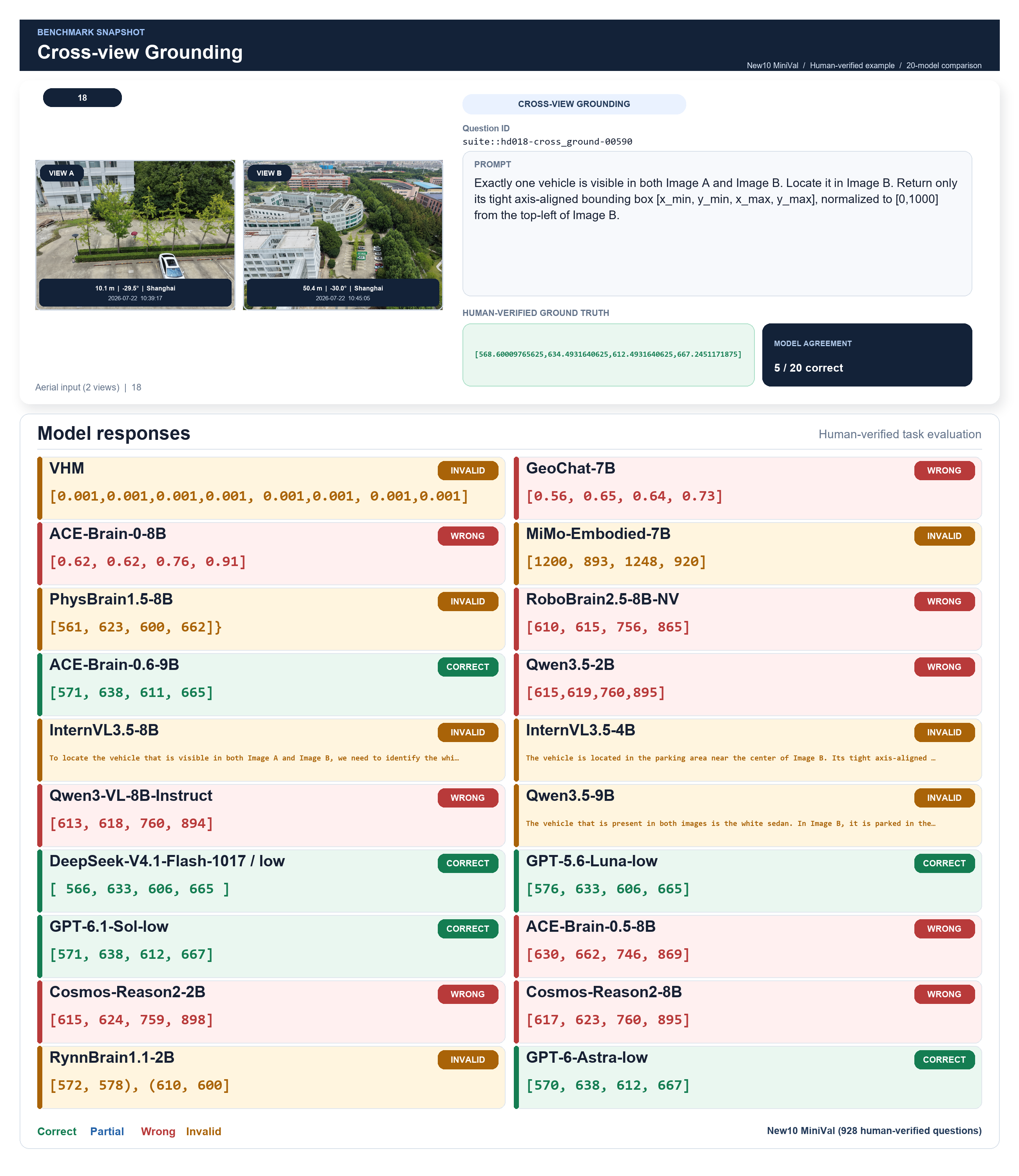}
\taskdemocard{RI: Instance re-identification}{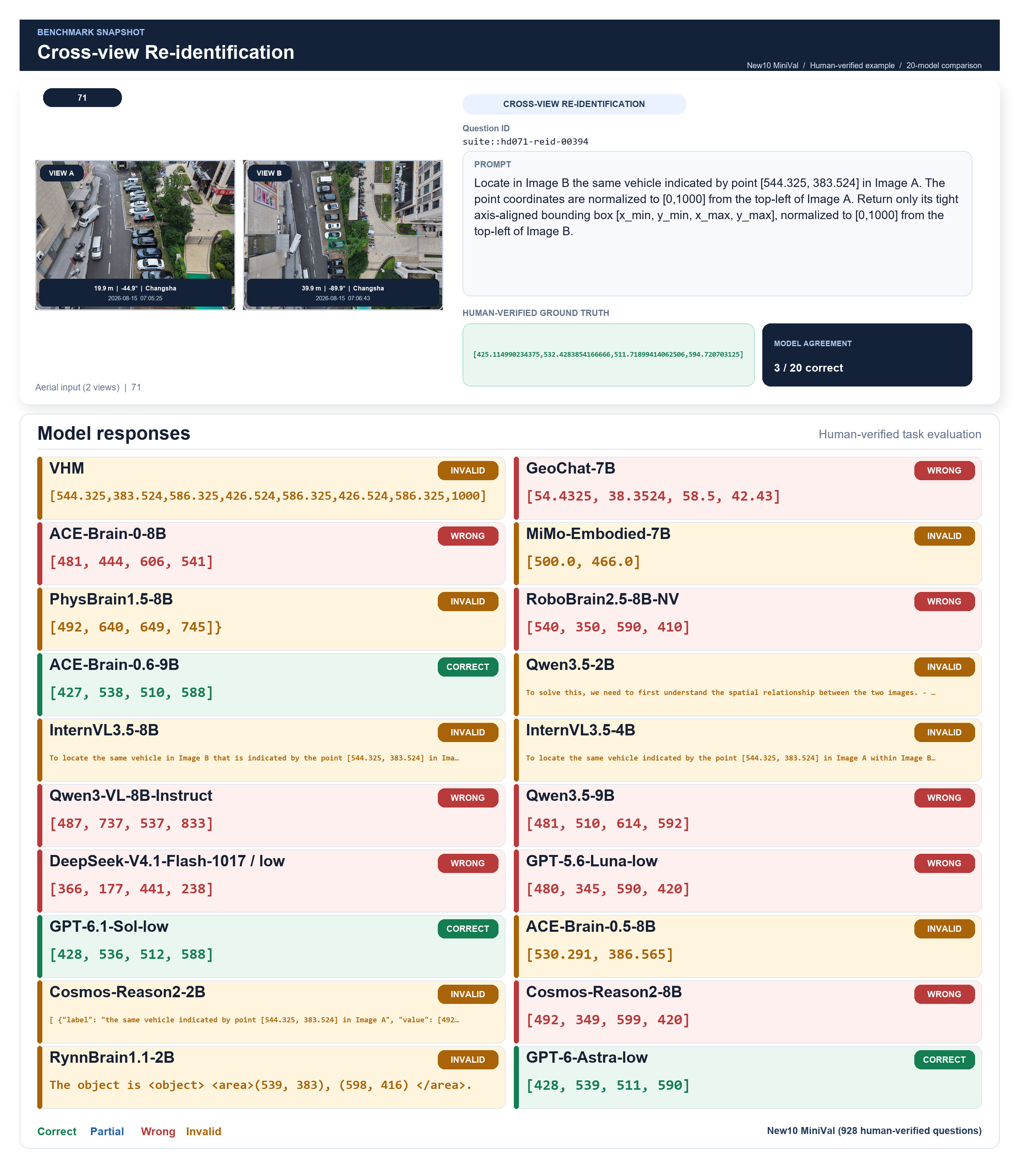}
\taskdemocard{CP: Camera height and pitch}{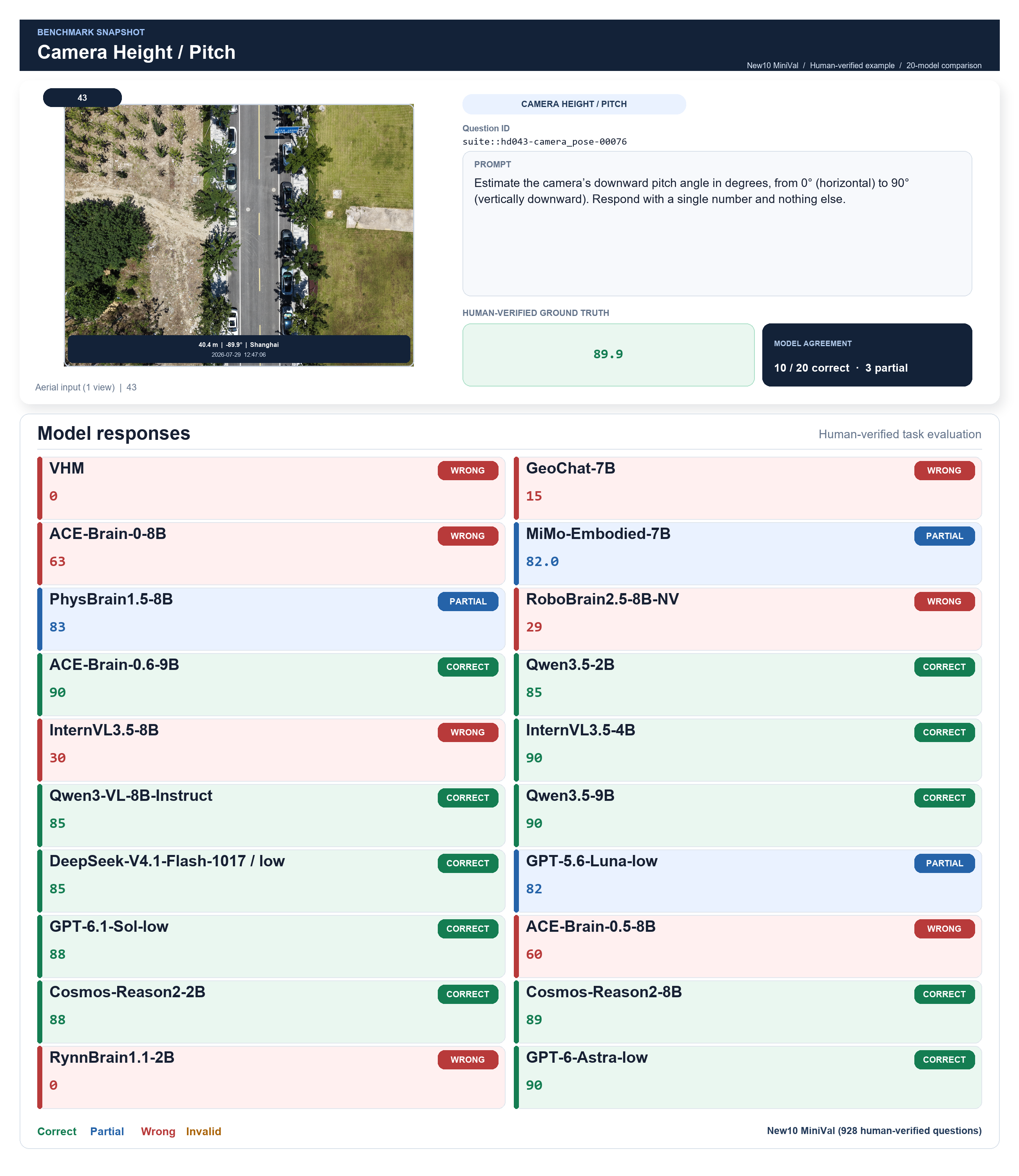}
\taskdemocard{PC: Vehicle-centred polar coordinates}{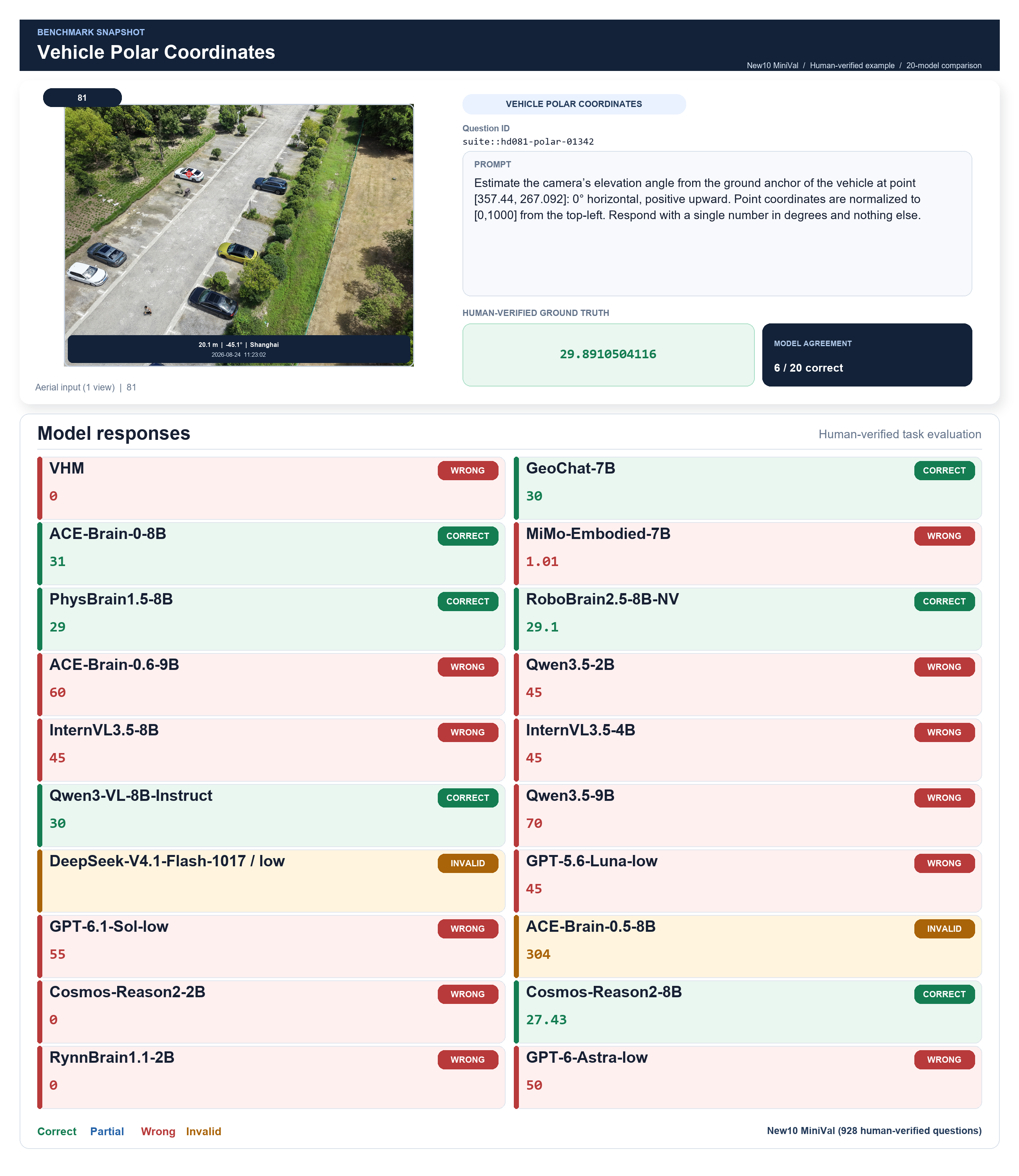}
\taskdemocard{CL: Camera location}{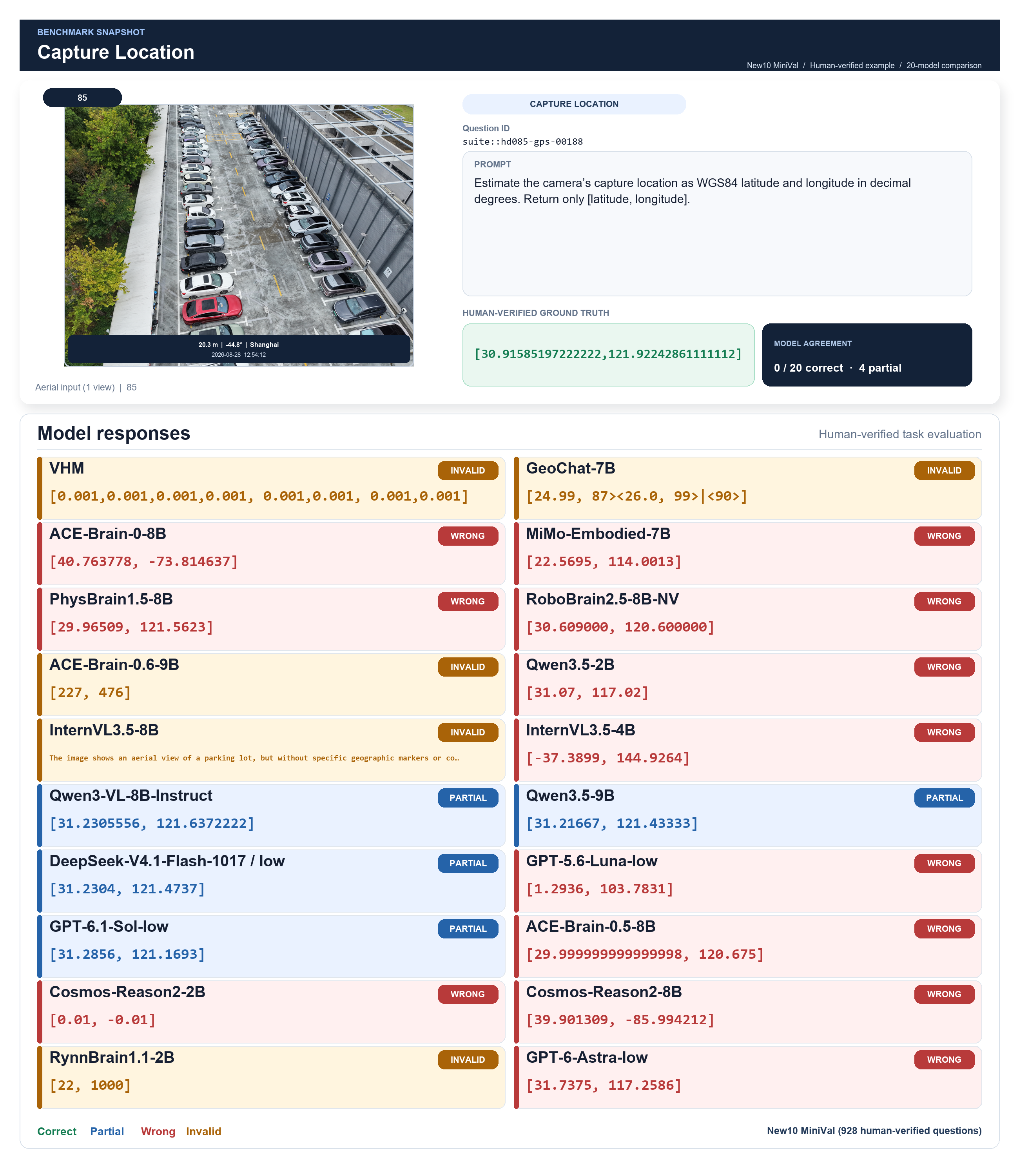}
\taskdemocard{CH: Capture hour}{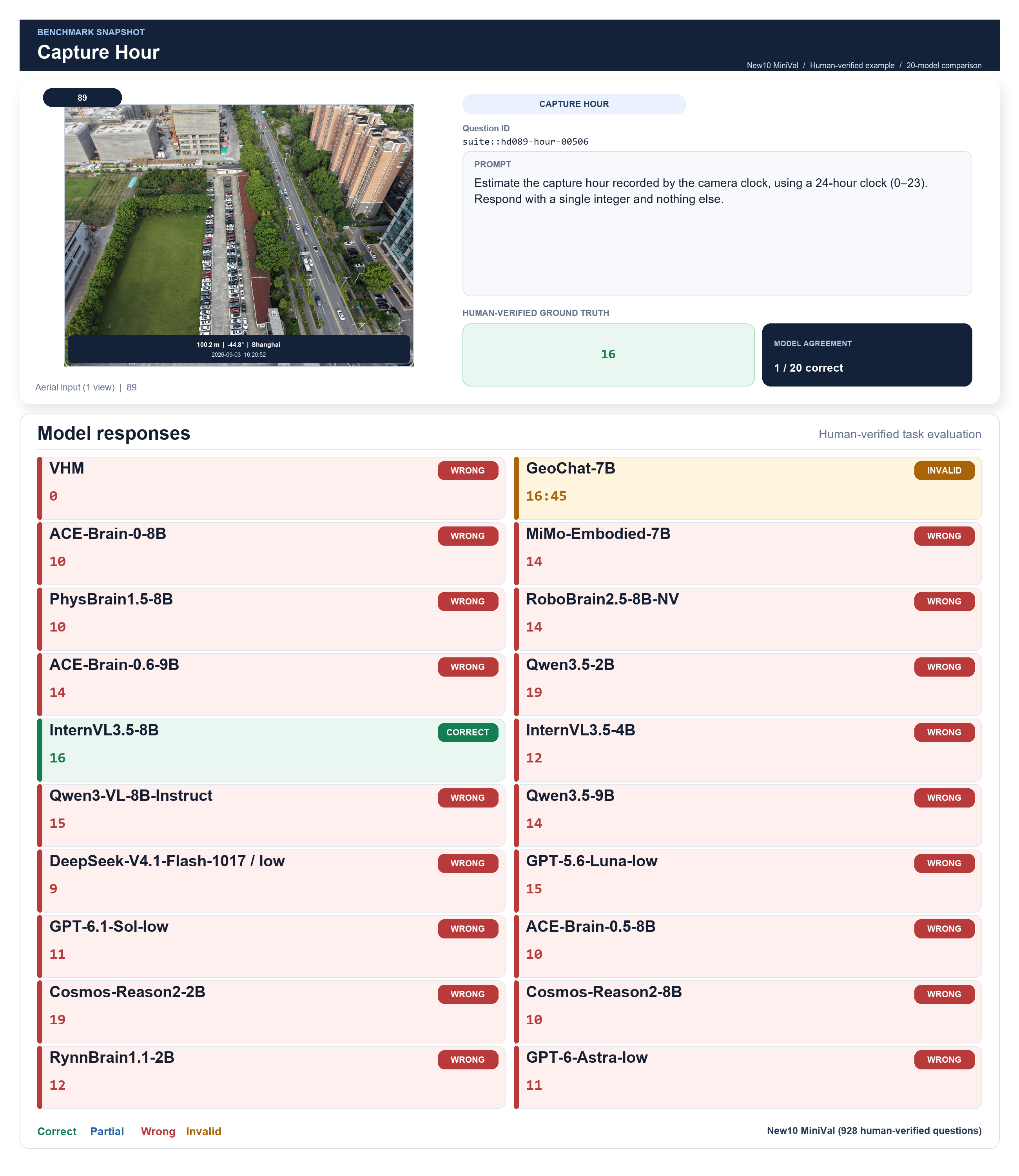}
\twocolumn

\end{document}